\pdfoutput=1

\documentclass[11pt]{article}

\usepackage{natbib}
\setcitestyle{numbers,square}
\usepackage{CUSTOM}

\usepackage{times}
\usepackage{latexsym}
\usepackage[T1]{fontenc}
\usepackage[utf8]{inputenc}
\usepackage{microtype}
\usepackage{inconsolata}
\usepackage{algorithm}
\usepackage{algorithmic}
\usepackage{amsmath}
\usepackage{graphicx}
\usepackage{multirow}
\usepackage{subcaption}
\usepackage{booktabs}
\usepackage{hyperref}
\usepackage{array} 
\usepackage{makecell}
\usepackage{enumitem}
\usepackage{longtable}
\usepackage{supertabular}
\usepackage{amssymb}
\usepackage{float}

\graphicspath{ {./figure} }

\title{PropNet: a White-Box and Human-Like Network for Sentence Representation}

\author{Fei Yang \\
         CogBeauty Lab \\
         yftadyz@163.com}

\begin{document}
\maketitle
\begin{abstract}

Transformer-based embedding methods have dominated the field of sentence representation in recent years. Although they have achieved remarkable performance on NLP missions, such as semantic textual similarity (STS) tasks, their black-box nature and large-data-driven training style have raised concerns, including issues related to bias, trust, and safety. Many efforts have been made to improve the interpretability of embedding models, but these problems have not been fundamentally resolved. To achieve inherent interpretability, we propose a purely white-box and human-like sentence representation network, PropNet. Inspired by findings from cognitive science, PropNet constructs a hierarchical network based on the propositions contained in a sentence. While experiments indicate that PropNet has a significant gap compared to state-of-the-art (SOTA) embedding models in STS tasks, case studies reveal substantial room for improvement. Additionally, PropNet enables us to analyze and understand the human cognitive processes underlying STS benchmarks.

\end{abstract}

\section{Introduction}

Sentence representation research has been carried out for over half a century, evolving from regular expressions \cite{1985Generalized}, statistical encoding methods \cite{harris54, Katz:87, 1968Automatic, jones72astatistical} to neural network embeddings \cite{mikolov2013efficient, bojanowski2016enriching, joulin2016bagtricksefficienttext, joulin2016fasttextzip, devlin2018bert, Reimers2019SentenceBERTSE, Gao2021SimCSESC, Li2024AoEAE}. In recent years, embedding models have dominated the research community, and also have achieved significant success in industry applications. However, their black-box nature and reliance on large-scale data-driven training raise concerns about bias \citep{conf/nips/BolukbasiCZSK16, Caliskan_2017, Brunet2018Understanding, Nadeem2021StereoSet}, trust \citep{Ribeiro2016why, conf/acl/RibeiroWGS20}, safety \citep{conf/naacl/BelinkovG19, conf/emnlp/PerezHSCRAGMI22} and other issues \citep{Bender2021On}. Researchers have attempted to improve the interpretability of embedding models \citep{conf/emnlp/JainBMMW18, conf/ijcnlp/LiaoLHL20} and remove their biases \citep{conf/naacl/GonenG19, conf/ltedi/GiraZL22}, but these problems remain fundamentally unsolved. Moreover, in the process of training models, massive amounts of manually labeled data have been generated. However, researchers typically focus solely on the training value of these labeled data, neglecting the cognitive values that exist within them. These cognitive values hold significant potential for enhancing human self-understanding.

In the cognitive science community, the question of how our brains represent a sentence has drawn a great deal of research interest \cite{Kintsch1974TheRO, Anderson1974RetrievalOP, alba1983is, ericsson1995longterm, 1998Linguistic, 2007The, Jeffrey2009Where}. When humans read and comprehend a sentence, they decompose it into propositions. A proposition is a simple sentence with no more than one verb. The meanings of these propositions are stored in long-term memory as a hierarchical network, which can be recalled for use in downstream tasks \citep{Kintsch1978TowardAM, Collins1969RetrievalTF}. The brain's verbal short-term memory possesses a specialized subsystem. This subsystem is responsible for splitting propositions and handling the syntactic features of each proposition as the eyes sweep across the text \citep{1999Verbal}. Moreover, the cognitive load in sentence comprehension is primarily attributed to syntactic complexity, such as the number of verbs, rather than superficial length, such as the addition of adjectives and other modifiers \citep{1998Linguistic}. This implies proposition is a processing unit for sentence comprehension. Inspired by these findings, in this work, we propose a purely white-box and human-like sentence representation network named PropNet. 

\begin{table*}[h!]
\centering
\begin{tabular}{p{0.3\textwidth}|p{0.126\textwidth}|p{0.15\textwidth}|p{0.1\textwidth}|p{0.1\textwidth}}
\hline
\textbf{Method} & \textbf{No Feature Engineering} & \textbf{Semantic Preservation} & \textbf{White\-Box} & \textbf{Human\-Like} \\ \hline
Regular Expressions & $\times$ & $\times$ & $\checkmark$ & $\times$ \\ \hline
Statistical Encoding & $\checkmark$ & $\times$ & $\checkmark$ & $\times$ \\ \hline
Neural Network Embedding & $\checkmark$ & $\checkmark$ & $\times$ & $\times$ \\ \hline
CISM & $\checkmark$ & $\times$ & $\checkmark$ & $\checkmark$ \\ \hline
PropNet & $\checkmark$ & $\checkmark$ & $\checkmark$ & $\checkmark$ \\ \hline
\end{tabular}
\caption{Comparison of Sentence Representation Methods}
\label{tab:method_comparison}
\end{table*}

PropNet constructs a hierarchical network based on the
propositions comprising a sentence. The generation of PropNet involves four phases: splitting, parsing, representing and merging. The first two phases employ a mechanism analogous to short-term memory, splitting a sentence into propositions and parsing key tokens from each proposition according to specific dimensions. The dimensions include \texttt{Action}, \texttt{Subject}, \texttt{Object}, \texttt{Where}, \texttt{Auxiliary\_Object}, \texttt{Goal}, \texttt{Reason}, and \texttt{Source}. The last two phases function akin to long-term memory, representing a proposition as a hierarchical network and merging all networks into a unified network to express the complete meaning of the original sentence. Figure~\ref{fig:ss1} provides an example of PropNet.

PropNet is purely white-box since it is completely interpretable and transparent. It is also human-like because it has a hierarchy network structure to save the meaning in a sentence. With respect to semantics, unlike embedding models which focus more on the relationships between words and their contextual information, PropNet focuses on the true happenings in the world as expressed by a sentence. It achieves this by analyzing the core components of propositions, particularly the eight dimensions related to verbs. The principles behind PropNet's design are very natural, without specially crafted artificial features. The comparison of PropNet and other representation methods is summarized in Table~\ref{tab:method_comparison}.

In semantic textual similarity (STS) tasks, given a pair of sentences, humans are required to assign a similarity score according to specific instructions. Cognitive processes play a key role in bridging sentences/instructions and annotation results. PropNet enables us to study this cognitive process. Experiments show that differences in actions, subjects or objects are a sufficient condition for humans to confirm an inequivalent proposition pair.

We compare PropNet with representative embedding methods on STS tasks. Although
experiments imply that PropNet has a large gap compared to state-of-the-art (SOTA) embedding models, case studies reveal significant potential for enhancement.

\section{Relative Work}

Text representation approaches could date back to rule-based regular expressions for search engines and question-answering systems \cite{1985Generalized}. Feature engineering is implemented heavily to develop expressions for pattern matching. This approach is inherently explainable, as the expressions consist of readable strings. However, it is pretty rigid and shallow, without a deep understanding of semantics. 

Statistical encoding methods compute word frequencies and represent text as discrete or continuous vectors. Typical methods include One Hot Embedding (OHE), Bag of Words (BOW) \cite{harris54}, N-grams \cite{Katz:87},  Normalized Term Frequency (NTF) \cite{1968Automatic} and TF-IDF \cite{jones72astatistical}. These methods are straightforward to implement since no feature engineering is required. They are also explainable, as each number in the representation vectors has a clear meaning. However, since word frequency cannot handle semantic issues, like synonyms and antonyms, or word relationships, these methods only provide shallow semantic analysis.

Neural network embedding approaches represent words as vectors in a continuous vector space. Representative methods include Word2Vec \cite{mikolov2013efficient}, FastText \cite{bojanowski2016enriching, joulin2016bagtricksefficienttext, joulin2016fasttextzip}, Glove \cite{2014Glove}, BERT \cite{devlin2018bert}, SBERT \citep{Reimers2019SentenceBERTSE}, SimCSE \citep{Gao2021SimCSESC} and AoE \citep{Li2024AoEAE}. They train neural networks on an extremely large corpus, using different Language Modeling (LM) tasks as the objective. Since embeddings are automatically computed by backpropagation, feature engineering is completely eliminated. In addition, the embeddings provide deep semantic understanding of the text, as they capture the semantic aspects of language due to LM tasks as the training goal. 

Embedding methods are considered as black-box algorithms because the embedding vectors are generally unexplainable. This characteristic surreptitiously encodes social biases contained in the training corpus \citep{conf/nips/BolukbasiCZSK16, Caliskan_2017, Brunet2018Understanding, Nadeem2021StereoSet}, and also raises concerns about trust \citep{Ribeiro2016why, conf/acl/RibeiroWGS20}, safety \citep{conf/naacl/BelinkovG19, conf/emnlp/PerezHSCRAGMI22} and other issues \citep{Bender2021On}. Numerous efforts have been made to improve interpretability \citep{conf/emnlp/JainBMMW18, conf/ijcnlp/LiaoLHL20} and remove biases \citep{conf/naacl/GonenG19, conf/ltedi/GiraZL22}, but these problems have not been fundamentally solved. 

All the aforementioned methods are not human-like, as their representation formats are either strings or vectors. Cognitive-Inspired Syntactic Models (CISM), such as Tree Adjoining Grammar \citep{joshi1975tree}, Combinatory Categorial Grammar \citep{steedman1987combinatory}, and Dependency Grammar \citep{hays1964dependency}, are based on principles of human language comprehension and production. These models use interpretable tree structures to represent sentences but primarily focus on syntactic organization. PropNet, which employs Dependency Grammar for parsing, enhances semantic representation through instance nodes and stamp nodes. Unlike CISM, PropNet uses a network structure rather than a tree structure to represent sentences. Table~\ref{tab:method_comparison} summarizes the characteristics of these methods.

\section{Methods}


\begin{figure*}[h!]
  \includegraphics[width=1\textwidth]{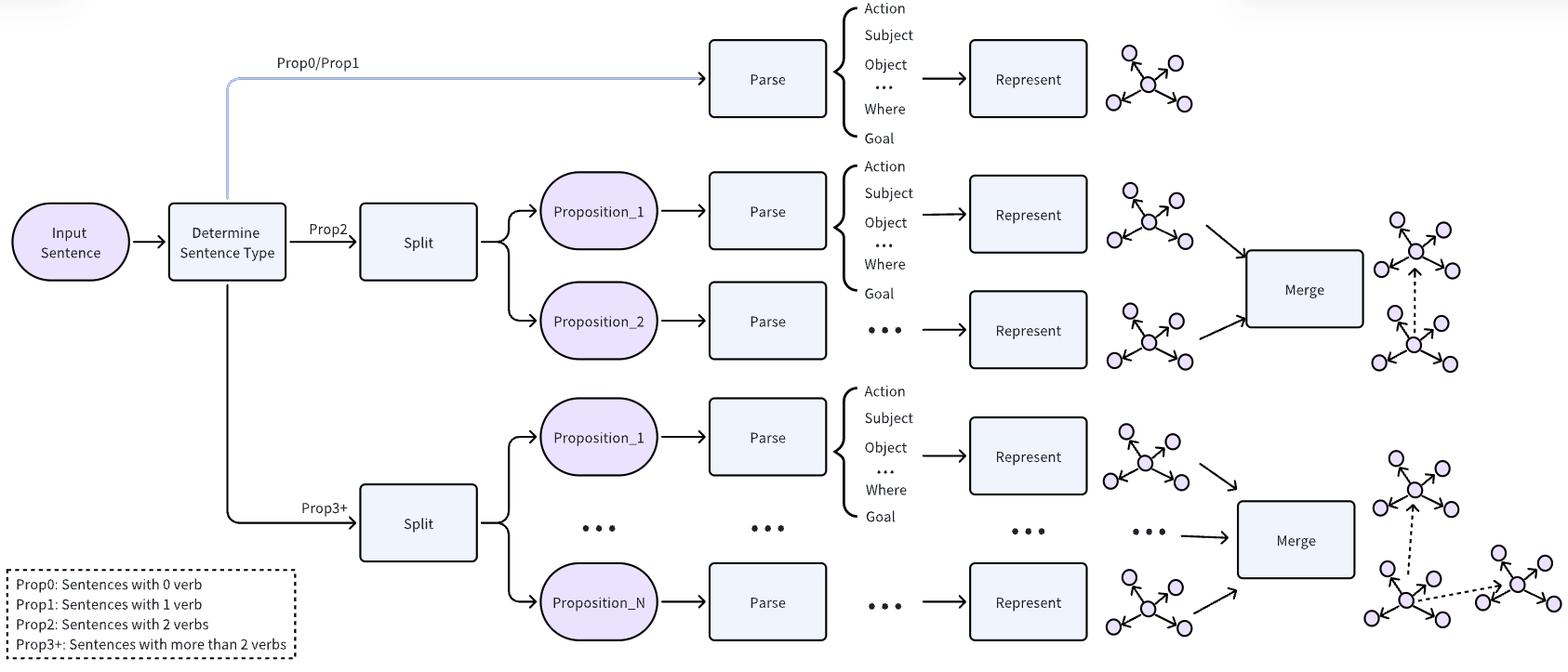}
  \caption{The framework of building PropNet. The type of the input sentence is calculated before splitting. After merging, all networks are integrated into one network as the PropNet of the input sentence. Note that for types \texttt{Prop0} and \texttt{Prop1} the splitting and merging phases are omitted.}
  \label{fig:framework}
\end{figure*}

\begin{table*}[h!]
\centering
\small 
\renewcommand{\arraystretch}{1.5}
\begin{tabular}{@{}>{\raggedright\arraybackslash}p{2.5cm}>{\raggedright\arraybackslash}p{2.8cm}>{\raggedright\arraybackslash}p{9cm}@{}}
\toprule
\textbf{Clause Type} & \textbf{Primary Rule} & \textbf{Example} \\ \midrule
Conjunct & token.dep\_ == \texttt{conj} & I am singing and \textbf{playing} a guitar. -> (I am singing, I be playing a guitar) \\
\midrule
Clausal Complement & token.dep\_ == \texttt{ccomp} & She thinks this \textbf{is} a good idea. -> (she thinks \texttt{identifier\_ccomp}, this is a good idea) \\
\midrule
Adnominal Clause & token.dep\_ ==  \texttt{acl} & A cat \textbf{sitting} on sand looks up at the camera. -> (a cat looks up at the camera, cat be sitting on sand) \\
\midrule
Prepositional Complement & token.dep\_ == \texttt{pcomp} & He goes to school by \textbf{taking} a bus. -> (he goes to school by \texttt{identifier\_pcomp}, he be taking a bus) \\
\midrule
Adverbial Clause & token.dep\_ == \texttt{advcl} & Although it is \textbf{raining}, we will go to a garden. -> (we will go to a garden, it is raining) \\
\midrule
Open Clausal Complement & token.dep\_ == \texttt{xcomp} & He is going to \textbf{swim}. -> (he is going to \texttt{xcomp\_identifier}, he swim) \\
\midrule
Subject Clause & token.dep\_ == \texttt{csubj} & \textbf{Adding} aspirin to the water could kill the plant. -> (\texttt{identifier\_csubj} could kill the plant, adding aspirin to the water) \\
\midrule
Relative Clause & token.dep\_ == \texttt{relcl} & They like the person who \textbf{lives} in the street. -> (They like the person, person lives in the street) \\ 
\bottomrule
\end{tabular}
\caption{Splitting rules for a \texttt{Prop2} sentence according to its clause types. The condition \texttt{token.pos\_ in (\texttt{VERB},\texttt{AUX})} is also required for all rules. We omit it to improve the clarity of the table. Note that token.pos\_ and token.dep\_ are the POS tagging and dependency parsing results from spaCy. The examples follow the format: Input Sentence -> (Main Proposition, Subordinate Proposition). The bold text in an example is the extracted verb token for the subordinate clause. Note that \texttt{identifier\_ccomp}, \texttt{identifier\_pcomp}, \texttt{xcomp\_identifier} and \texttt{identifier\_csubj} serve as placeholders for the relevant clauses.  
} 
\label{tab:splitting_rules_cs2v}
\end{table*}


To build a PropNet of a sentence, we first split the sentence into propositions (Section~\ref{method:splitting}). Each proposition corresponds to a verb in the sentence. Then propositions are parsed to extract core tokens with respect to dimensions \texttt{Action}, \texttt{Subject}, \texttt{Object}, \texttt{Where}, \texttt{Auxiliary\_Object}, \texttt{Goal}, \texttt{Reason}, and \texttt{Source} (Section~\ref{method:parse}). After that, each proposition is represented by a hierarchical network (Section~\ref{method:represent}). At last, all networks are merged into one network as the PropNet of the input sentence (Section~\ref{method:merge}). Figure~\ref{fig:framework} illustrates the framework of constructing PropNet. We introduce a method for comparing two PropNets, thereby unlocking the potential of the resulting PropNets for STS tasks (Section~\ref{sec: diff_vec}).     

The following symbols are used to represent different types of sentences based on the number of verbs they contain:

\begin{itemize}
    \item \texttt{Prop0}: Sentences with 0 verbs, e.g., ``A ball on the ground.''
    \item \texttt{Prop1}: Sentences with 1 verb, e.g., ``A boy is playing football.''
    \item \texttt{Prop2}: Sentences with 2 verbs, e.g., ``A boy is playing football and a girl is dancing.''
    \item \texttt{Prop3+}: Sentences with more than 2 verbs, e.g., ``A man is sitting in a chair, wearing a cloak, and holding a stick.''
\end{itemize}

Here the term "verb" encompasses linking verbs. For instance, in the sentences "a boy is in a park" or "a girl is happy," the word "is" is treated as a verb, and both sentences are categorized as \texttt{Prop1}. In the rest of this work, a proposition refers to a \texttt{Prop1} or \texttt{Prop0} sentence.

We introduce the following notations to classify sentence pairs based on sentence types:

\begin{itemize}
    \item \texttt{P1-} denotes the set of pairs where both sentences are \texttt{Prop0} or \texttt{Prop1}.
    \item \texttt{P2} represents the set of pairs that contain at least one \texttt{Prop2}, excluding pairs with any \texttt{Prop3}. For instance, a \texttt{Prop1} and a \texttt{Prop2} sentence, or two \texttt{Prop2} sentences. 
    \item \texttt{P3+} indicates the set of pairs that includes at least one \texttt{Prop3} sentence.
\end{itemize}

\subsection{Splitting}
\label{method:splitting}

\begin{figure*}[h!]
  \includegraphics[width=1\textwidth]{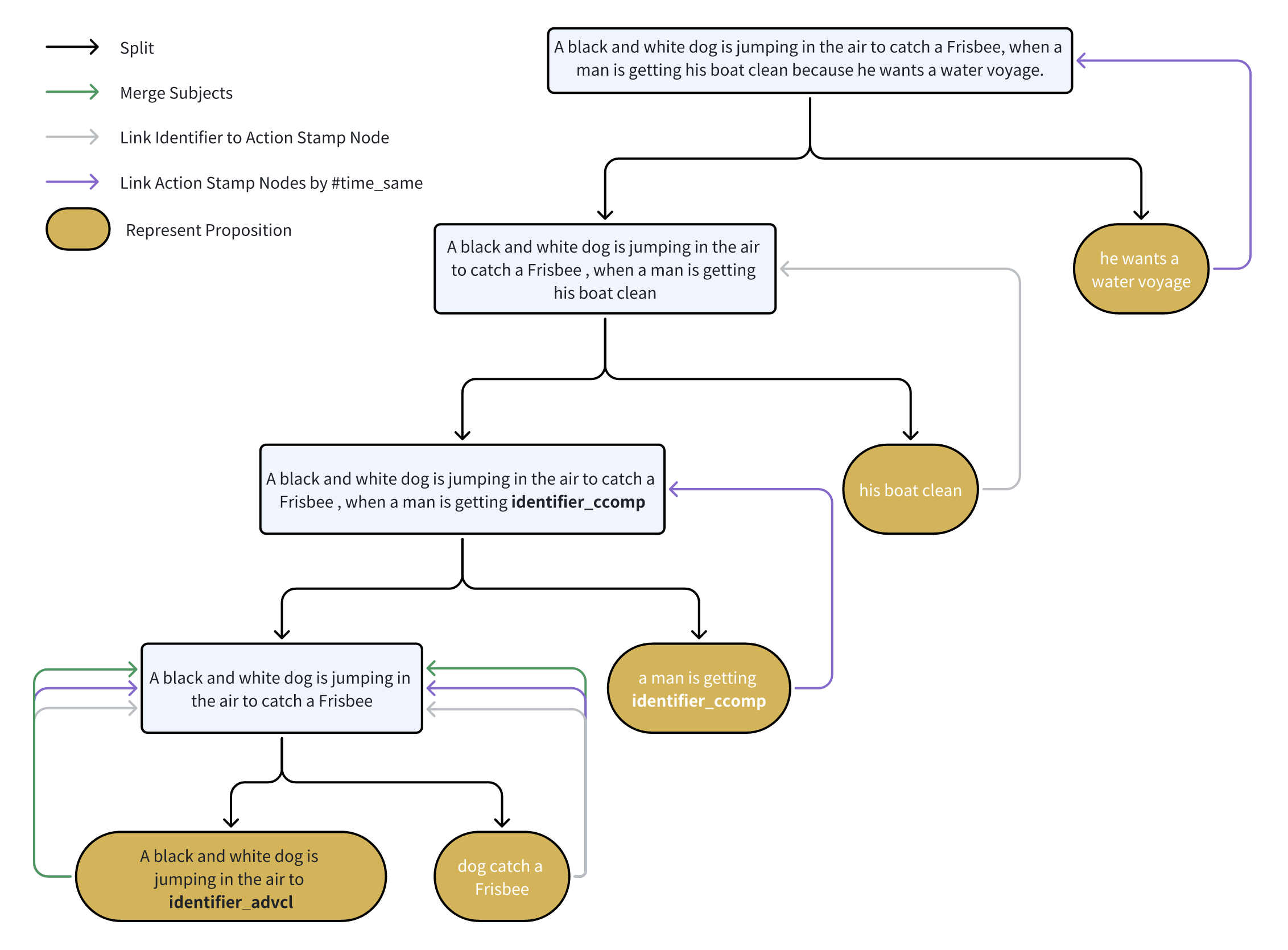}
  \caption{Splitting a \texttt{Prop3+} sentence and merging its proposition representations by backtracking. At each splitting step, the method for splitting \texttt{Prop2} is called. The extracted proposition invokes the representation component to build a hierarchical network. During the upward traversal, the same strategies for merging \texttt{P2}---denoted by green, grey, and purple arrows---are employed to integrate all networks into a unified network. Note that \texttt{identifier\_advcl} is a placeholder for an adverbial clause.}
  \label{fig:split_and_merge_cs_plus3v}
\end{figure*}

For \texttt{Prop0} and \texttt{Prop1}, splitting is unnecessary. At this stage, the input sentence itself serves as the output proposition. For \texttt{Prop2}, we split the input sentence into two \texttt{Prop1} propositions, a main proposition (main clause) and a subordinate proposition (subordinate clause). Table~\ref{tab:splitting_rules_cs2v} presents the splitting rules for different clause types. These rules extract the verb token of the subordinate clause. Then the subordinate clause is generated by iterating over the verb token's subtree provided by spaCy parsing tool \footnote{For more details, see the spaCy documentation: \url{https://spacy.io/usage/linguistic-features/\#pos-tagging}}. For instance, in the \texttt{Prop2} sentence ``She thinks this is a good idea'', the splitting rules identify ``is'' as the verb token for the subordinate clause. The token ``is'' owns a subtree ``this is a good idea'' by dependency parsing. 


The generated subordinate clause is then replaced by a clause identifier in the input sentence for five clause types \texttt{Clausal Complement}, \texttt{Prepositional Complement}, \texttt{Open Clausal Complement}, \texttt{Subject Clause} and \texttt{Adverbial Clause}. Their identifiers are notated as \texttt{identifier\_ccomp}, \texttt{identifier\_pcomp}, \texttt{xcomp\_identifier}, \texttt{identifier\_csubj} and \texttt{identifier\_advcl}, respectively. For other clause types, the subordinate clause is replaced by an empty string. When the replacement is completed, the rest of the input sentence becomes the main proposition. Table~\ref{tab:splitting_rules_cs2v} presents examples for each clause type. Note that \texttt{identifier\_advcl} is applied only when the clause is preceded by the word ``to'' that indicates the purpose of an action. Figure~\ref{fig:split_and_merge_cs_plus3v} provides an example for this specific situation. 

Clause identifiers are crucial for merging proposition representations into a unified network. In the above example, the main proposition is ``She thinks \texttt{identifier\_ccomp}'' where \texttt{identifier\_ccomp} would point to ``this is a good idea'' in the merging phase.

If the subordinate clause is non-finite, the subject of the main clause is spliced to it, forming a complete subordinate proposition. For instance, in ``she wants to dance'', the subordinate proposition ``she dance'' uses the subject of the main proposition ``she wants xcomp\_identifier''. If the subordinate clause is finite, no subject insertion is required since the clause already contains its own subject. The splitting rules for relative clause type are more intricate and are detailed in Appendix~\ref{app:propnet_relative_clause}.

Situation becomes significantly more complex for \texttt{Prop3+}. We employ a backtracking algorithm to decompose the input sentence into \texttt{Prop1} propositions by constructing a binary tree, as shown in Figure~\ref{fig:split_and_merge_cs_plus3v}. The splitting method for \texttt{Prop2} is invoked during each splitting step to extract a subordinate proposition \texttt{Prop1}, serving as the leaf node of the binary tree. Then the leaf node invokes the representation component to construct a hierarchical network, which is explained in Section~\ref{method:represent}. The leaf nodes indicate the order of the propositions contained in the long input sentence from left to right. In Figure~\ref{fig:split_and_merge_cs_plus3v}, the input sentence has an ordered proposition list [``A black and white dog is jumping in the air to identifier\_advcl'', ``dog catch a Frisbee'', ``a man is getting identifier\_ccomp'', ``his boat clean'', ``he wants a water voyage''] according to the leaf nodes from left to right. This order is necessary when we compare two \texttt{Prop3+} sentences for similarity, which is explained in Section~\ref{sec: diff_vec}.

\subsection{Parsing}
\label{method:parse}

\begin{table*}[h!]
	\footnotesize
	\centering 
    \small 
    \renewcommand{\arraystretch}{1.5}
	{
		\begin{tabular}{p{0.08\textwidth}|p{0.28\textwidth}|p{0.30\textwidth}|p{0.22\textwidth}}
			\toprule 
			\textbf{Dimension} & \textbf{Explanation} & \textbf{Primary Rule} & \textbf{Example}\\
            
			\midrule
			Action & An action described by the verb in a proposition. & token.pos\_ is \texttt{VERB} & The man is \textbf{riding} a horse. \\
            
			Subject * & An entity or concept that performs the action. & token.dep\_ is \texttt{nsubj} & The \textbf{man} is riding a horse. \\
            
            Object * & An entity or concept that receives the action. & token.dep\_ in (\texttt{dobj}, \texttt{nsubjpass}) & The man is riding a \textbf{horse}. \\		
            \midrule
            
            Where *	&The place where the action occurs, or the time when it occurs.	& preposition of token in ``on'', ``in'', ``inside'', ``outside'', ``at'', \ldots)	&The polar bear is sliding on the \textbf{snow}.	\\
            
            Aux\_Obj *	&The instrument or tool that is used to perform the action. Or indirect object which is the recipient, beneficiary, or target of an action.  	& preposition of token in (``with'', ``by'', ``about'', ``as'', \ldots)	&A person is cutting mushrooms with a \textbf{knife}.		\\	
            
            Goal *	&The intended outcome or objective that the subject aims to achieve by performing the action.	& preposition of token in (``into'', ``to'', ``onto'', ``towards'',\ldots)	&A man pours oil into a \textbf{pot}.\\
            
            Reason *	&The cause or motivation behind the action.	& preposition of token in (``for'', ``due'')	&The teacher praised her for her excellent \textbf{work}.		\\
            
            Source *	&The origin or starting point from which the action arises. 	&preposition of token in (``from'')	&She learned the news from her \textbf{friend}.		\\	
            \midrule
            
            Attribute	&The appearance, state, nature, or features of an entity or a concept.	&token.dep\_ in (\texttt{amod}, \texttt{nmod}, \texttt{compound}) 	&A \textbf{young} girl is using sign language.		\\														
            Part\_of	&One entity is a component or piece of another larger entity.	& An entity token of another entity token; token.dep\_ is \texttt{poss}	& Two men are packing suitcases into the trunk of a \textbf{car}.		\\															
			\bottomrule
		\end{tabular}
	}
    \caption{ Dimensions of parsing a \texttt{Prop1}/\texttt{Prop0} sentence. Dimensions \texttt{Attribute} and \texttt{Part\_of} are designed for an entity or concept. Others are related to the action in a proposition. Primary rules of extracting tokens are explained with examples. Note that token.pos\_ and token.dep\_ are the POS tagging and dependency parsing results from spaCy. The bold text in an example is the extracted token for its dimension. Dimensions marked with a ``*'' also follow this parsing rule \texttt{token.pos\_ in (\texttt{NOUN}, \texttt{PRON}, \texttt{PROPN}, \texttt{NUM})} which is omited in the table for brevity.
	} \label{tab:parsing_rules}	
\end{table*}

This phase extracts key tokens from a proposition with respect to specific dimensions. The intuition of designing these dimensions is that we require they can describe the meaning of a single action that happens in the real world, which is the key to understand the meaning of a complex sentence consisting of a series of actions. Therefore, eight critical dimensions related to the verb of a proposition are considered: \texttt{Action}, \texttt{Subject}, \texttt{Object}, \texttt{Where}, \texttt{Auxiliary\_Object}, \texttt{Goal}, \texttt{Reason}, \texttt{Source}. The extracted words under each dimension are either describing a physical entity like ``car'' or an abstractive concept like ``brand''. For these words, two additional dimensions \texttt{Attribute} and \texttt{Part\_of} are considered for the descriptions of them. Explanations for these dimensions and extracting primary rules \footnote{Because of the variety of language, even a \texttt{Prop1} sentence can have an extremely flexible structure. Therefore, minor rules are handcrafted for extraction which can be seen in the codes.} with examples are listed in Table~\ref{tab:parsing_rules}. The prepositions used by primary rules in the table are omitted for brevity. See Appendix~\ref{app:propnet_parse} for a detailed list of these prepositions. Extraction is implemented by resorting to spaCy part-of-speech tagging and dependency parsing tools \footnote{For more details, see the spaCy documentation: \url{https://spacy.io/usage/linguistic-features/\#pos-tagging}}.

\subsection{Representing}
\label{method:represent}

\begin{figure*}[h!]
  \includegraphics[width=1\textwidth]{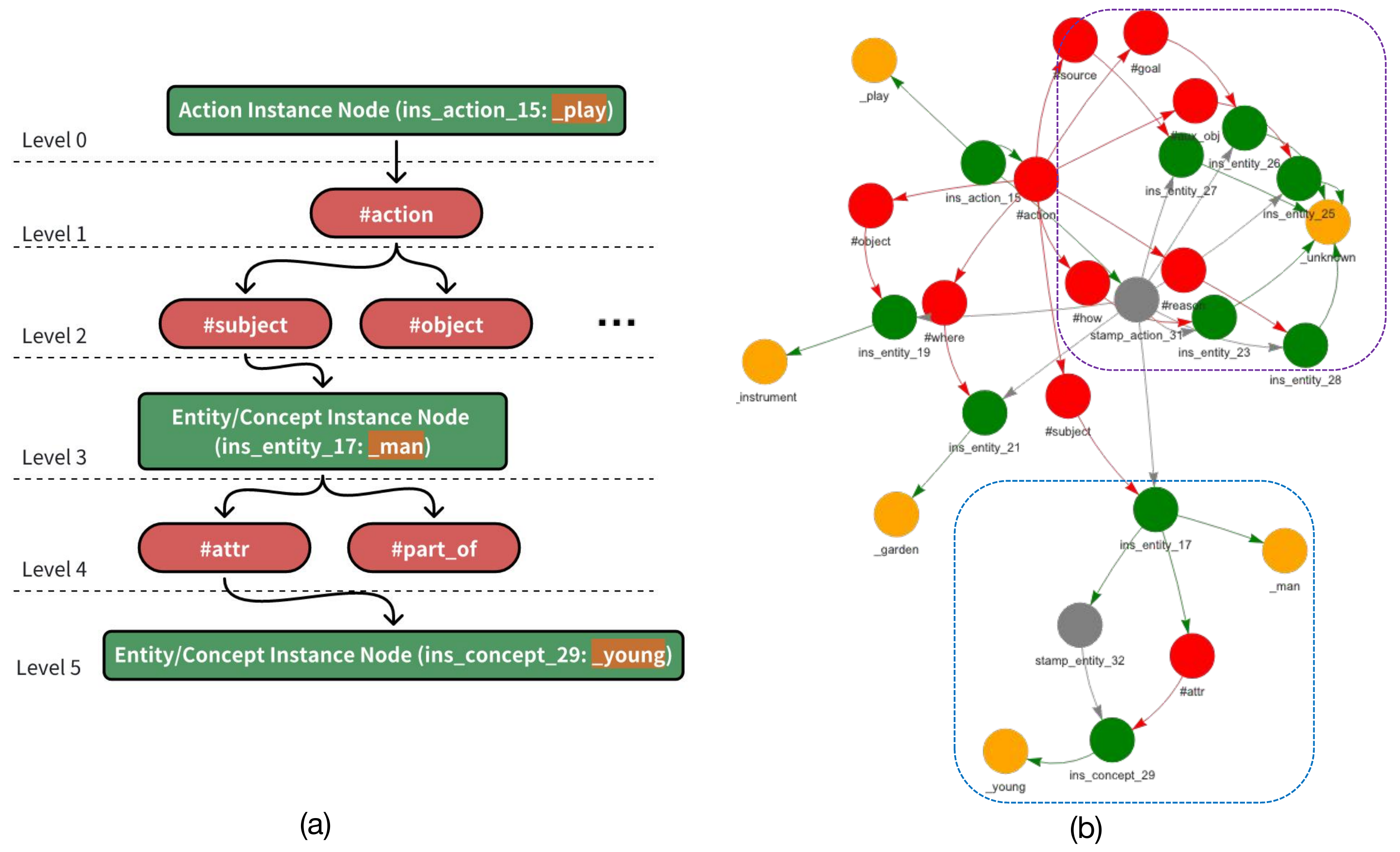}
  \center
  \caption{(a) Six-level hierarchical structure of PropNet. (b) An example of representing the sentence ``A young man is playing an instrument in the garden''. Evolutionary nodes, developmental nodes, instance nodes and stamp nodes are labeled as red, orange, green and grey colors respectively.}
  \label{fig:ss1}
\end{figure*}

As brain's long-term memory stores information as a hierarchical network \citep{Collins1969RetrievalTF, Anderson1974RetrievalOP, Kintsch1974TheRO}, PropNet represents a proposition using a six-level hierarchical network. Figure~\ref{fig:ss1} (b) provides an example of how the sentence ``A young man is playing an instrument in the garden'' is represented. The network comprises three types of nodes: evolutionary nodes, developmental nodes and instance nodes which are represented as red, orange and green colors, respectively. In designing the network, we incorporate ideas similar to Nature Design and Nurture Beliefs , which are outlined in \citep{Yang2024AutomaticEO}. Evolutionary nodes correspond to Nature Design, which reflects human nature and is derived from heredity. They are denoted with a prefix ``\#'', like \texttt{\#action}. Each dimension in the parsing phase has a corresponding evolutionary node. Developmental nodes correspond to Nurture Beliefs, which are learned from experience, and are named with a prefix ``\_'', such as \texttt{\_young}. Each word in a sentence corresponds to a developmental node. 

An evolutionary node cannot directly establish a connection with a developmental node. This connection is fulfilled by an instance node. Each evolutionary node points to at least one instance node. Each instance node can point to only one developmental node. They have the following three types:

\begin{itemize}
    \item Entity instance nodes are linked to developmental nodes whose lexical content has a hypernym ``physical entity'', a relationship that can be ascertained using WordNet \citep{miller1995wordnet}. Entity instance nodes are denoted as \texttt{ins\_entity\_id}; for instance, \texttt{ins\_entity\_17}, which is linked to \texttt{\_man}.
    \item Concept instance nodes are linked to developmental nodes whose lexical content is not ``physical entity'', such as adjectives, pronouns, and numbers. These nodes are denoted as \texttt{ins\_concept\_id}; for example, \texttt{ins\_concept\_29}, which is connected to \texttt{\_young}.
    \item Action instance nodes are linked to developmental nodes that contain verbal lexical content and are labeled as \texttt{ins\_action\_id}. As an example, \texttt{ins\_action\_15} refers to \texttt{\_play}.
\end{itemize}

Figure~\ref{fig:ss1} (a) provides a more lucid depiction of the six-level hierarchical structure of PropNet. An action instance node functions as the root and triggers \texttt{\#action}. It also points to a developmental node \texttt{\_play}. This action instance node is unique because there is only one verb in a proposition. Edges from \texttt{\#action} connect evolutionary nodes \texttt{\#subject}, \texttt{\#object}, \texttt{\#where}, \texttt{\#aux\_obj}, \texttt{\#goal}, \texttt{\#source}, and \texttt{\#reason}, each of which corresponds to a parsing dimension described in Section~\ref{method:parse}. These nodes will establish connections with instance nodes that are created for the words parsed from each dimension. For example, \texttt{\#subject} points to \texttt{ins\_entity\_17}, which has a link with \texttt{\_man}. At level 4, these instance nodes activate two evolutionary nodes, \texttt{\#attr} and \texttt{\#part\_of}, which align with the parsing dimensions \texttt{Attribute} and \texttt{Part\_of} respectively. Similarly, \texttt{\#attr} and \texttt{\#part\_of} establish links with instance nodes created for the words parsed under \texttt{Attribute} and \texttt{Part\_of}. For instance, \texttt{ins\_entity\_17} points to \texttt{\#attr} and \texttt{\#attr} points to \texttt{\_young}.

\begin{table*}[h!]
\centering
\small
\renewcommand{\arraystretch}{1.2}
\begin{tabular}{@{}p{2.5cm}p{6cm}p{6cm}@{}}
\toprule
\textbf{Strategy} & \textbf{Rule} & \textbf{Example} \\ \midrule
Merge Subjects & Merge the instance nodes with the same subject into a single instance node. & A \textbf{cat} sitting on sand looks up at the camera. $\rightarrow$ (a \textbf{cat} looks up at the camera, \textbf{cat} be sitting on sand) [Merge instance nodes of \textbf{cat}] \\

Link Clause Identifier to Action Stamp Node (ASN) & Establish links between the instance node of \texttt{identifier\_ccomp}, \texttt{identifier\_pcomp}, \texttt{xcomp\_identifier}, \texttt{identifier\_advcl} and \texttt{identifier\_csubj} in the MAIN and the ASN in the SUB. (MAIN: main proposition; SUB: subordinate proposition) & She thinks this is a good idea. $\rightarrow$ (she thinks identifier\_ccomp, this is a good idea) [Link the instance node of \textbf{identifier\_ccomp} to the ASN invoked by \textbf{is}] \\
Link Action Stamp Nodes (ASNs) by Time Node & 

\begin{itemize}[leftmargin=*,nosep,after=\strut]
    \item \textbf{MAIN Before SUB}: Link the ASN of MAIN to \texttt{\#time\_before}, and then link \texttt{\#time\_before} to the ASN of SUB;
    \item \textbf{MAIN After SUB}: Link the ASN of SUB to \texttt{\#time\_before}, and then link \texttt{\#time\_before} to the ASN of MAIN;
    \item \textbf{MAIN and SUB Occur Simultaneously}: Link \texttt{\#time\_same} to the ASNs of MAIN and SUB.
\end{itemize} &

The boy washes his hands before he has lunch. $\rightarrow$ (the boy washes his hands, he has lunch) [ASN invoked by \textbf{wash} -> \texttt{\#time\_before} -> ASN invoked by \textbf{has}] \\ \bottomrule
\end{tabular}
\caption{Merging Strategies for A \texttt{Prop2} Sentence}
\label{tab:merge_2v_cs}
\end{table*}

If during the parsing phase, no words are associated with a dimension, its corresponding evolutionary node will invoke an instance node that points to a default developmental node, \texttt{\_unknown}. This ensures the network remains complete even when some information is missing. In Figure~\ref{fig:ss1} (b), only \texttt{\#subject}, \texttt{\#object} and \texttt{\#where} have valid instance nodes, which point to \texttt{\_man}, \texttt{\_instrument} and \texttt{\_garden}, correspondingly. All other evolutionary nodes pointed to by \texttt{\#action}  are connected to instance nodes that point to \texttt{\_unknown}, which is marked by a purple box in the figure.

There's a fourth type of nodes, called stamp nodes, which can be distinguished by their grey color in Figure~\ref{fig:ss1} (b). The primary role of stamp nodes is to distinguish different parts of the network representing a complex sentence that consists of several propositions. An action stamp node is activated by an action instance node and named as \texttt{stamp\_action\_id}. It points to all entity or concept instance nodes associated with that action, thereby grouping them together and differentiating them from the entity or concept instance nodes of other actions. 

An entity stamp node is triggered by an entity instance node and named as \texttt{stamp\_entity\_id}. It points to all entity or concept instance nodes that describe the corresponding entity instance node. As shown by the blue box in the figure, \texttt{stamp\_entity\_32} points to \texttt{ins\_concept\_29}, aggregating the modifier ``young'' to the entity ``man''.

\subsection{Merging}
\label{method:merge}
When the main proposition and the subordinate proposition of a \texttt{Prop2} sentence are represented separately, we need to combine them to convey a complete and coherent meaning. We put forward three merging strategies, which are enumerated in Table~\ref{tab:merge_2v_cs}. If the subordinate proposition is a relative clause, a different merging strategy is adopted, which is presented in Appendix~\ref{app:propnet_relative_clause}. 

For a \texttt{Prop3+} sentence, the merging process corresponds to the upward backtracking process of splitting, illustrated in Figure~\ref{fig:split_and_merge_cs_plus3v}. The same strategies are applied as for \texttt{Prop2}. Appendix~\ref{app:propnet_cases} demonstrates the merging result of the sentence example in Figure~\ref{fig:split_and_merge_cs_plus3v}.

\subsection{Comparison of Two PropNets}
\label{sec: diff_vec}

Given the PropNets of a pair of sentences, a natural consideration is how to compare the differences between the two networks. However, since the PropNet of a complex sentence can be extensive, containing a large number of nodes and edges, directly comparing two PropNets poses a significant challenge due to computational complexity. A more viable computational approach involves comparing the disparities in the representations of the corresponding propositions between the two sentences, and then aggregating these disparities into an overall signature difference. Thus, we initially introduce the method for comparing the differences between two propositions, which is previously presented as pair type \texttt{P1-}. Next, based on this foundation, we discuss the difference-construction methods for more complex pair types \texttt{P2} and \texttt{P3+}.

\begin{figure*}[h]
  \includegraphics[width=1\textwidth]{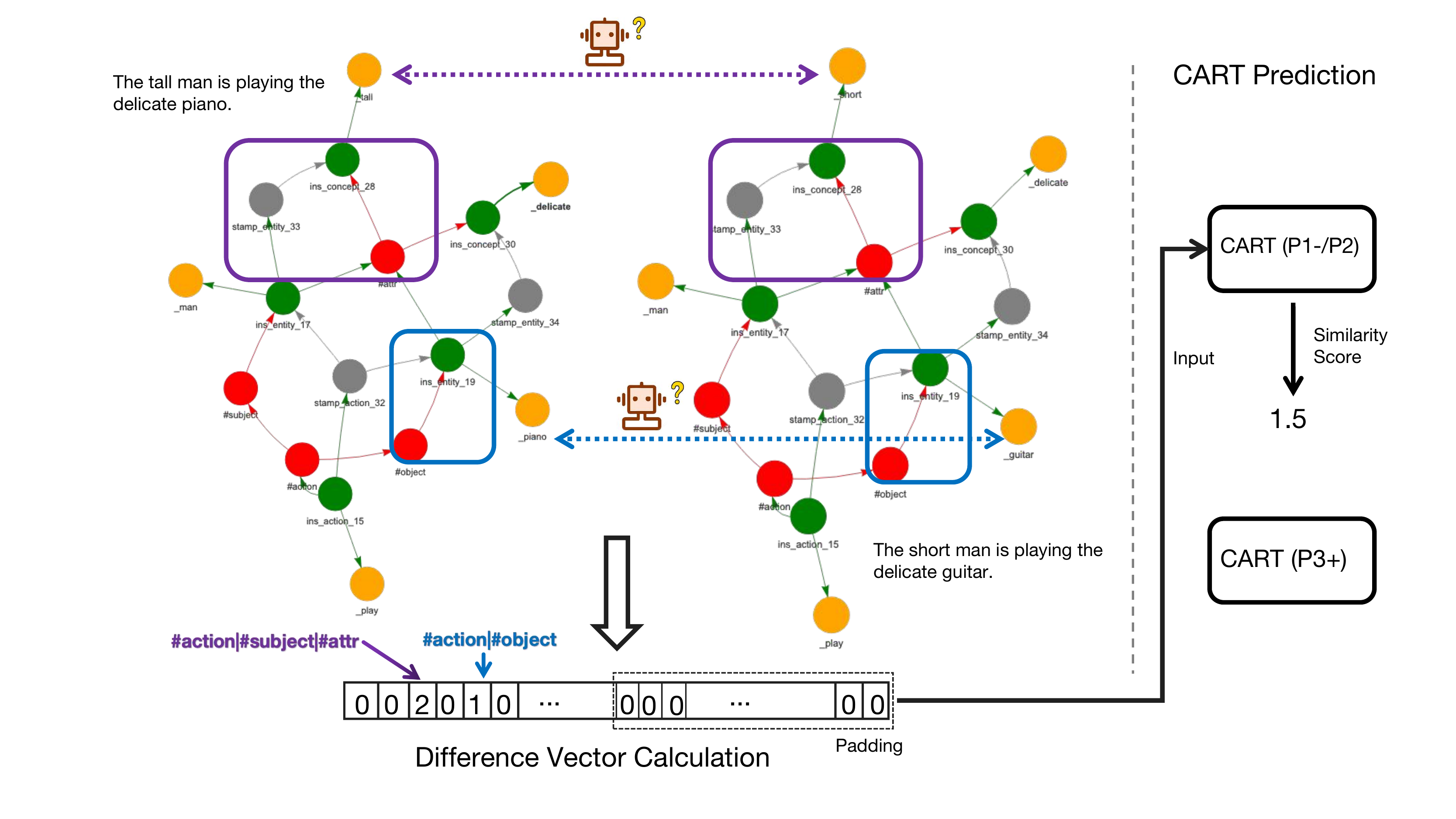}
  \centering
  \caption{Framework of the comparison module. This module consists of two parts: difference vector computation and CART prediction. It is exemplified by the pair ``The tall man is playing the delicate piano'' and ``The short man is playing the delicate guitar''. They differ at \texttt{\#action|\#subject|\#attr} and \texttt{\#action|\#object}, which are marked with elements 2 and 1 at the corresponding positions in the difference vector. The codes 0, 1, 2 represent ``identical'', ``similar'' and ``different'', respectively. Note that the difference vector for \texttt{P1-} is padded with 0, doubling its size, before it enters the corresponding CART model.}
  \label{fig:compare_module}
\end{figure*}

\subsubsection{Pair Type \texttt{P1-}}
\label{pair_type_p1}

The comparison dimensions are formed by the Cartesian product of \{\texttt{\#subject}, \texttt{\#object}, \texttt{\#where}, \texttt{\#aux\_obj}, \texttt{\#goal}, \texttt{\#source}, \texttt{\#reason}\} and \{$1$, \texttt{\#attr}, \texttt{\#part\_of}\}, further extended by \{\texttt{\#action}, \texttt{\#action|\#subject|\#where}, \texttt{\#action|\#object|\#where}\}. For example, the dimension \texttt{\#action|\#subject|\#attr} represents the attribute of an entity or concept that serves as the subject within the action. The reason of using the names of evolutionary nodes is this way of naming can reveals the route to locate the corresponding entity or concept instance nodes within the graph of PropNet. The explanations of all these 24 dimensions are provided in Appendix~\ref{sec:compare_dim}. 

By comparing the developmental nodes indicated by the dimensional instance nodes, we obtain a similarity code for each dimension. Concatenating these codes in sequence produces a difference vector $\mathbf{v_1}$. The similarity code primarily takes three values: 0 for identical, 1 for similar, 2 for different. A large language model (LLM) acts as the knowledge base to compute the similarity degree between two words. This strategy is word-level based, ignoring context-dependent meanings, e.g. polysemous words. Identification of the gloss that a word uses given a context is left in future work. Details on LLM choice and its prompts are explained in Appendix~\ref{app:llm_lexical_similarity}. Additional similarity codes, and examples for each code are provided in Appendix~\ref{app:similar_code}.

Figure~\ref{fig:compare_module} illustrates how to calculate a difference vector given the representations of two sentences ``The tall man is playing the delicate piano'' and ``The short man is playing the delicate guitar''. Purple boxes identify the concept instance nodes under \texttt{\#action|\#subject|\#attr} with two different developmental nodes \texttt{\_tall} and \texttt{\_short}. Blue boxes identify the entity instance nodes under \texttt{\#action|\#object} with two different developmental nodes \texttt{\_piano} and \texttt{\_guitar}. The instance nodes of other dimensions have exactly the same developmental nodes. Since \texttt{\_tall} and \texttt{\_short} are judged as different by LLM, \texttt{\#action|\#subject|\#attr} contributes a code of 2 to the difference vector. As \texttt{\_piano} and \texttt{\_guitar} is judged as similar by LLM, \texttt{\#action|\#object} contributes a code of 1 to the difference vector.

\subsubsection{Pair Type \texttt{P2}}
Assume the first sentence \texttt{s1} of the pair being compared can be split into two propositions, \texttt{pp1} and \texttt{pp2}. If the second sentence \texttt{s2} is \texttt{Prop0} or \texttt{Prop1}, the following alignment process is applied. If two propositions share identical subjects and actions, they are considered to have a significant overlap. This criterion can be evaluated by examining the specific elements \texttt{\#action} and \texttt{\#action|\#subject} of the difference vector. For a fair comparison, we align \texttt{s2} to the proposition that exhibits a significant overlap with it. Subsequently, we utilize the difference vector associated with this alignment. This vector is then padded with 2 until it attains a length twice that of the original, resulting in the final output $\mathbf{v_2}$. Padding with 2 implies \texttt{s2} lacks a corresponding component to compare with the other proposition of \texttt{s1}.

We illustrate the alignment process by comparing ``I like apple'' and ``She likes orange and I like banana''. ``She likes orange and I like banana'' is split into ``She likes orange'' and ``I like banana''. According to the order of appearance, we would compare ``She likes orange'' and ``I like apple'', which is not a fair comparison. By implementing the alignment process, ``I like banana'' and ``I like apple'' are compared, while ``She likes orange'' has no counterpart for comparison.

If the second sentence \texttt{s2} is \texttt{Prop2}, a similar alignment process is applied, but without padding. For example, in the pair ``I like apple and she likes pineapple'' and ``She likes orange and I like banana'', the proposition ``I like apple'' aligns with ``I like banana'', and ``she likes pineapple'' aligns with ``she likes orange''.

\subsubsection{Pair Type \texttt{P3+}}
As a \texttt{Prop3+} sentence can contain multiple propositions, we restrict our consideration to its first four propositions in the order of the leaf nodes in its splitting binary tree from left to right. Given a \texttt{P3+} pair, suppose the first sentence $s_1$ and the second sentence $s_2$ have ordered proposition lists $L_1 = [p_1, p_2, p_3, p_4]$ and $L_2 = [q_1, q_2, q_3, q_4]$ respectively. Note that $p_2, p_3, p_4$ from $s_1$ and $q_4$ from $s_2$ might be empty strings. We calculate the Proposition Overlap Degree (POD) as the number of overlapping words between two propositions. $L_2$ is aligned with $L_1$ according to POD by the following steps:




\begin{enumerate}[label=(\arabic*)]
    \item The PODs between $p_1$ and each one of $L_2$ are calculated. Suppose $q_1^{(1)}$ from $L_2$ has the largest POD which is greater than a threshold value 1, then $q_1^{(1)}$ is moved to the front of $L_2$. Other elements in $L_2$ keep their orders invariant. After this rearrangement, $L_2$ is notated as $L_2 = [q_1^{(1)}, q_2^{(1)}, q_3^{(1)}, q_4^{(1)}]$. If the largest POD is no greater than 1, then $L_2$ keeps its original order, which means $L_2 = [q_1^{(1)}, q_2^{(1)}, q_3^{(1)}, q_4^{(1)}] = [q_1, q_2, q_3, q_4]$.

    \item Suppose $L_2^{(1)}=[q_2^{(1)}, q_3^{(1)}, q_4^{(1)}]$. Compute the PODs between $p_2$ and each one of $L_2^{(1)}$. Similarly, check if the largest POD is greater than 1, and rearrange $L_2^{(1)}$ to $[q_2^{(2)}, q_3^{(2)}, q_4^{(2)}]$.

    \item Suppose $L_2^{(2)} = [q_3^{(2)}, q_4^{(2)}]$. Compute the PODs between $p_3$ and each one of $L_2^{(2)}$. Check if the largest POD is greater than 1, and rearrange $L_2^{(2)}$ to $[q_3^{(3)}, q_4^{(3)}]$. Finally, $s_2$ has a new ordered proposition list $[q_1^{(1)}, q_2^{(2)}, q_3^{(3)}, q_4^{(3)}]$ which is aligned to $s_1$ according to POD.

\end{enumerate}

After alignment, there are four \texttt{P1-} pairs: $p_1$ and $q_1^{(1)}$, $p_2$ and $q_2^{(2)}$, $p_3$ and $q_3^{(3)}$, $p_4$ and $q_4^{(3)}$. We compute a difference vector $\mathbf{v_1}$ for each of them using previous computing method for a \texttt{P1-} pair. The final difference vector for this \texttt{P3+} pair is the concatenation of these four $\mathbf{v_1}$s. Therefore, the difference vector $\mathbf{v_3}$ for \texttt{P3+} has a size that is twice the length of $\mathbf{v_2}$ for \texttt{P2}. If two propositions are both empty, $\mathbf{v_1}$ is replaced by a vector of the same size, where all elements are 0. If one proposition is empty and the other one is not, $\mathbf{v_1}$ is replaced by a vector of the same size, where all elements are 2. 

Let's exemplify the above processes, considering the pair: ``Three men are jumping off a wall'' and ``Three young men run, jump, and kick off of a Coke machine''. The second sentence consists of three propositions: ``three young men run'', ``three young men jump'', and ``three young men kick off of a Coke machine''. After the alignment process, its proposition list is reordered to ``three young men jump'', ``three young men run'', ``three young men kick off of a Coke machine''. The first sentence is compared with ``three young men jump'', resulting a difference vector $\mathbf{v_1}$. Since ``three young men run'' and ``three young men kick off of a Coke machine''
have no counterparts in the first sentence, we concatenate $\mathbf{v_1}$ with two vectors of the same length as $\mathbf{v_1}$, padded with 2 to signify the two sentences are different in this part. Finally, the resulting vector is padded with 0 until it reaches the required length, indicating the two sentences are the same in this default part.

\section{Experiments}


\subsection{Comparison with Embedding Models}

\begin{table*}[h!]
\small
\centering
\begin{tabular}{lcccccccc}
\toprule
Model & STS12 & STS13 & STS14 & STS15 & STS16 & STS-B & SICK-R & Avg. \\

\midrule
InferSent-GloVe (2017)$\dagger$  & $52.86$ & $66.75$ & $62.15$ & $72.77$ & $66.87$ & $68.03$ & $65.65$ & $65.01$ \\
USE (2018)$\dagger$  & $64.49$ & $67.80$ & $64.61$ & $76.83$ & $73.18$ & $74.92$ & $76.69$ & $71.22$ \\
SBERT (2019) $\dagger$ & $70.97$ & $76.53$ & $73.19$ & $79.09$ & $74.30$ & $77.03$ & $72.91$ & $74.89$ \\
SimCSE-BERT (2021) & $68.40$ & $82.41$ & $74.38$ & $80.91$ & $78.56$ & $76.85$ & $72.23$ & $76.25$ \\
AoE-BERT (2024) & $75.26$ & $85.61$ & $80.64$ & $86.36$ & $82.51$ & $85.64$ & $80.99$ & $82.43$ \\

\midrule

PropNet & $\mathbf{47.01}$ & $\mathbf{36.33}$ & $\mathbf{40.74}$ & $\mathbf{34.54}$ & $\mathbf{12.06}$ & $\mathbf{47.41}$ & $\mathbf{55.35}$ & $\mathbf{39.06}$ \\ 

\bottomrule
\end{tabular}
\caption{Performance comparison with embedding models on STS tasks. We use Spearman's rank correlation coefficient ($\rho$) multiplied by 100 as evaluation metric. Results of the models with $\dagger$ are from \citep{Reimers2019SentenceBERTSE}, others are from their original papers. There is a significant gap between PropNet and the latest embedding models.
}
\label{table-main-sts-results}
\end{table*}

To assess the performance gap between PropNet and mainstream approaches, PropNet is compared with widely used embedding methods on semantic textual similarity (STS) tasks. 

\textbf{Baselines.} Given that our method is supervised, widely used supervised embedding methods are selected as baselines: InferSent \citep{2017Supervised}, USE \citep{Cer2018UniversalSE}, SBERT \citep{Reimers2019SentenceBERTSE}, SimCSE \citep{Gao2021SimCSESC}, and AoE \citep{Li2024AoEAE}.

\textbf{Datasets.} We conduct experiments on three benchmark datasets: STS 2012-2016 \citep{Agirre2012SemEval2012T6, Agirre2013SEM2S, Agirre2014SemEval2014T1, Agirre2015SemEval2015T2, Agirre2016SemEval2016T1}, STS-B \citep{Cer2017SemEval2017T1}, and Sick-R \citep{Marelli2014ASC}.

\textbf{Evaluation Metrics.} We employ Spearman's rank correlation coefficient ($\rho$) multiplied by 100 to measure the alignment between predicted and ground-truth scores. 

\textbf{Implementation Details. } The experiment employs a comparison module which consists of two components, (1) the calculation of the difference vector and (2) prediction using Classification and Regression Trees (CART) \citep{Breiman1984ClassificationAR}. Figure~\ref{fig:compare_module} presents the details of this module. After computing the difference vectors using the approach introduced in Section~\ref{sec: diff_vec} for all pairs within the benchmark datasets, two CART models are trained on these vectors: one for short pairs \texttt{P1-}/\texttt{P2} and another for long pairs \texttt{P3+}, because the sizes of the difference vectors are different. These models are denoted as \texttt{CART(P1-/P2)} and \texttt{CART(P3+)} respectively. We call the API provided by scikit-learn \footnote{API: \url{https://scikit-learn.org/stable/modules/generated/sklearn.tree.DecisionTreeRegressor.html}} for training. Since STS 2012-2016 and STS-B share the same rules for ground scores, training is implemented only on the train set of STS-B to learn these rules. Similarly, for SICK-R, two CART models are trained on its training set. The random seed is fixed as 0, and the minimum number of samples at a leaf node is set to 10. Further training details are appended in Appendix~\ref{app:cart_train_details}.

Note that the predictions for short pairs and long pairs in the test set are concatenated before computing the Spearman's correlation.

\textbf{Results. } In Table~\ref{table-main-sts-results}, a significant gap is observed between PropNet and the latest embedding model, AoE, which scores above 80. This indicates the efficacy of big-data-trained embeddings in comprehending text semantics. This table also demonstrates that embedding models have undergone notable development over the last ten years, with scores ranging from 65.01 to 82.43. Since PropNet marks the inception of a completely new research direction, offering purely white-box and human-like sentence representation, we are confident that PropNet will improve in the future, considering that there is substantial room for enhancement in its current version, which is presented in the next section.

If we only focus on the scores of PropNet across all benchmarks, we observe a significant deviation, ranging from 12.06 on STS 2016 to 55.35 on SICK-R. The genre of STS 2016 is mainly forum question/answer and news, while SICK-R is image/video description. This genre-score discrepancy raises the question of whether PropNet has different representation abilities across different genres or if the comparison module fails for a specific genre. We investigate this factor in detail in the next section.

\subsection{Investigation of Genres}

In this section, we examine the performance of PropNet across different genres. The number of verbs is also considered as an important influencing factor. We anticipate that predicting accurately using our comparison module becomes more challenging for complex sentences.

\textbf{Dataset.} We utilize the STS-B test set \footnote{Approximately 8.7\% of the STS-B records lack similarity scores due to spaCy dependency parsing errors. These records are excluded from the analysis.} along with its predicted scores for analysis. The test set is split into three parts according to its field \texttt{genre}, which has three values: \texttt{main-captions}, \texttt{main-news} and \texttt{main-forums}. They represent image/video description, news and forum question/answer, respectively. Appendix~\ref{app:genres} provides more detailed explanations of these genres.

Another splitting dimension is verb number. The test set is divided into three parts according to pair types \texttt{P1-}, \texttt{P2}, and \texttt{P3+}. Generally, from \texttt{P1-} to \texttt{P3+}, each sentence in a pair contains an increasing number of verbs. Through intersection operation, pairs restricted to a specific genre and verb number can be selected.

\textbf{Evaluation Metrics.} Spearman's rank correlation coefficient ($\rho$) multiplied by 100, is used to measure the alignment between predicted and ground truth scores for a specific subset of the STS-B test set. Moreover, the number of pairs in each subset, and their proportion relative to the entire test dataset are calculated to illustrate data distribution.

\begin{table*}[h]
\centering
\begin{tabular}{@{\hspace{5pt}}l@{\hspace{5pt}}c@{\hspace{5pt}}c@{}c@{\hspace{5pt}}c@{\hspace{5pt}}c@{}c@{\hspace{5pt}}c@{\hspace{5pt}}c@{}c@{\hspace{5pt}}c@{\hspace{5pt}}c}

	\multicolumn{1}{c}{\multirow{2}{*}{}} & \multicolumn{2}{c}{Total} & & \multicolumn{2}{c}{P1-} & & \multicolumn{2}{c}{P2} & &\multicolumn{2}{c}{P3+} \\
    
	\cmidrule{2-3} \cmidrule{5-6} \cmidrule{8-9} \cmidrule{11-12}
    
	& Count (\%) &  \  $\rho\times100$ & & Count (\%) & \ $\rho \times 100$& & Count (\%) &  \ $\rho \times 100$ & &  Count (\%) &  \ $\rho \times 100$\\
	\midrule
    
    \texttt{main-captions} & 616 (50\%)& 65.12 && 484 (39\%)& 70.41 & & 120 (10\%) & 43.15& &12 (1\%)&28.09 \\
    \texttt{main-news} & 411 (33\%)& 36.80 && 207 (17\%)&40.37 & & 124 (10\%) & 18.15& &80 (6\%) &20.22 \\
    \texttt{main-forums} & 212 (17\%)& 19.98 && 91 (7\%)& 43.53 & & 69 (6\%) & 26.70& &52 (4\%) &-0.04 \\
    Total & 1239 (100\%)&47.41 && 782 (63\%)& 59.76 & & 313 (25\%) & 26.11& &144 (12\%)&23.13 \\

	\bottomrule
\end{tabular}
\caption{Pair distributions and Spearman’s correlation scores corresponding to each genre and verb-number splitting. Count (\%) represents the number of pairs and their percentage of the total number. The results indicate that: (1) PropNet performs significantly better on \texttt{main-captions} than the other two genres; (2) As the number of verbs increases, Spearman scores decline significantly across all genres; (3) \texttt{main-news} and \texttt{main-forums} contain relatively more \texttt{P2} and \texttt{P3+} pairs than \texttt{main-caption}, implying these two genres have more complex sentence structures.}
\label{eval_error_anal}
\end{table*}

\textbf{Results.} As shown in Table~\ref{eval_error_anal}, PropNet scores 65.12 on \texttt{main-captions}, significantly higher than the 36.80 on \texttt{main-news} and 19.98 on \texttt{main-forums}. Even for the simplest pairs, \texttt{P1-}, PropNet attains only 40.37 on \texttt{main-news} and 43.53 on \texttt{main-forums}. This indicates that PropNet performs poorly on \texttt{main-news} and \texttt{main-forums}. The worst-performing pairs in \texttt{P1-}, whose prediction scores have the largest discrepancy with their ground scores, are analyzed to identify the underlying reasons. 


For \texttt{main-news}, the lack of social common knowledge contributes to the failure of the comparison module for \texttt{P1-}. For instance, it cannot accurately measure the similarity between the pair "bitcoin haul" and "bitcoins" without knowledge of the terms "bitcoin" and "haul". Another reason is that the evolutionary graph of PropNet is too coarse to handle detailed comparisons. For example, in the sentence "At Northwest Medical Center of Washington County in Springdale , one child is in serious condition", the phrase "in serious condition" is incorrectly regarded as a description of \texttt{\#where}. However, it should be linked to child's \texttt{\#state}, which is not included in current PropNet's evolutionary graph. 

For \texttt{main-forums}, PropNet currently lacks nodes to measure the likelihood of an action, such as modal verbs (e.g., "can", "should") and negation indicators (e.g., "not", "hardly"). For example, it fails in comparing ``You should do it'' and ``You can do it'', or ``It's not a good idea'' and ``It's a good question''. In addition, \texttt{main-forums} also suffers from the coarseness of the evolutionary graph. For instance, expressions like "It is up to you" cannot be represented properly. 

Regarding verb number, Table~\ref{eval_error_anal} indicates that as the number of verbs increases, Spearman scores across all genres decrease substantially. The worst-performing pairs in \texttt{P2} for \texttt{main-news} and \texttt{main-forums} are analyzed to identify the reason. 

For \texttt{main-news}, the use of clear and concise language in its writing style sometimes causes difficulties for dependency parsing, leading to failed representations. For instance, consider the sentence "Coroner: Whitney Houston died in bathtub." PropNet cannot properly represent "Coroner: " as "Coroner claims". 

For \texttt{main-forums}, sentences describing mental activities are often redundant, which complicates the alignment of propositions for computing the difference vector in the comparison module. For instance, in the pair "There are two things to consider" and "A couple things to consider", the phrase "there are" hinders proper alignment. A potential improvement could be simplifying "There are two things to consider" to "two things to consider" during comparison. 

All the above case studies imply that there is a substantial improvement room in PropNet. Another point worthy of note is that, for \texttt{main-news} and \texttt{main-forums}, the internal proportions of \texttt{P2} and \texttt{P3+} are 49.6\% (204/411) and 57.1\% (121/212) respectively, which are significantly higher than 21.4\% (132/616) of \texttt{main-captions}. This suggests that sentences in \texttt{main-news} and \texttt{main-forums} contain more complex structures.

\subsection{Human Cognitive Differences in STS Tasks}
\label{cog_diff_in_sts}

\begin{figure*}[h]
  \centering
  \includegraphics[width=2\columnwidth]{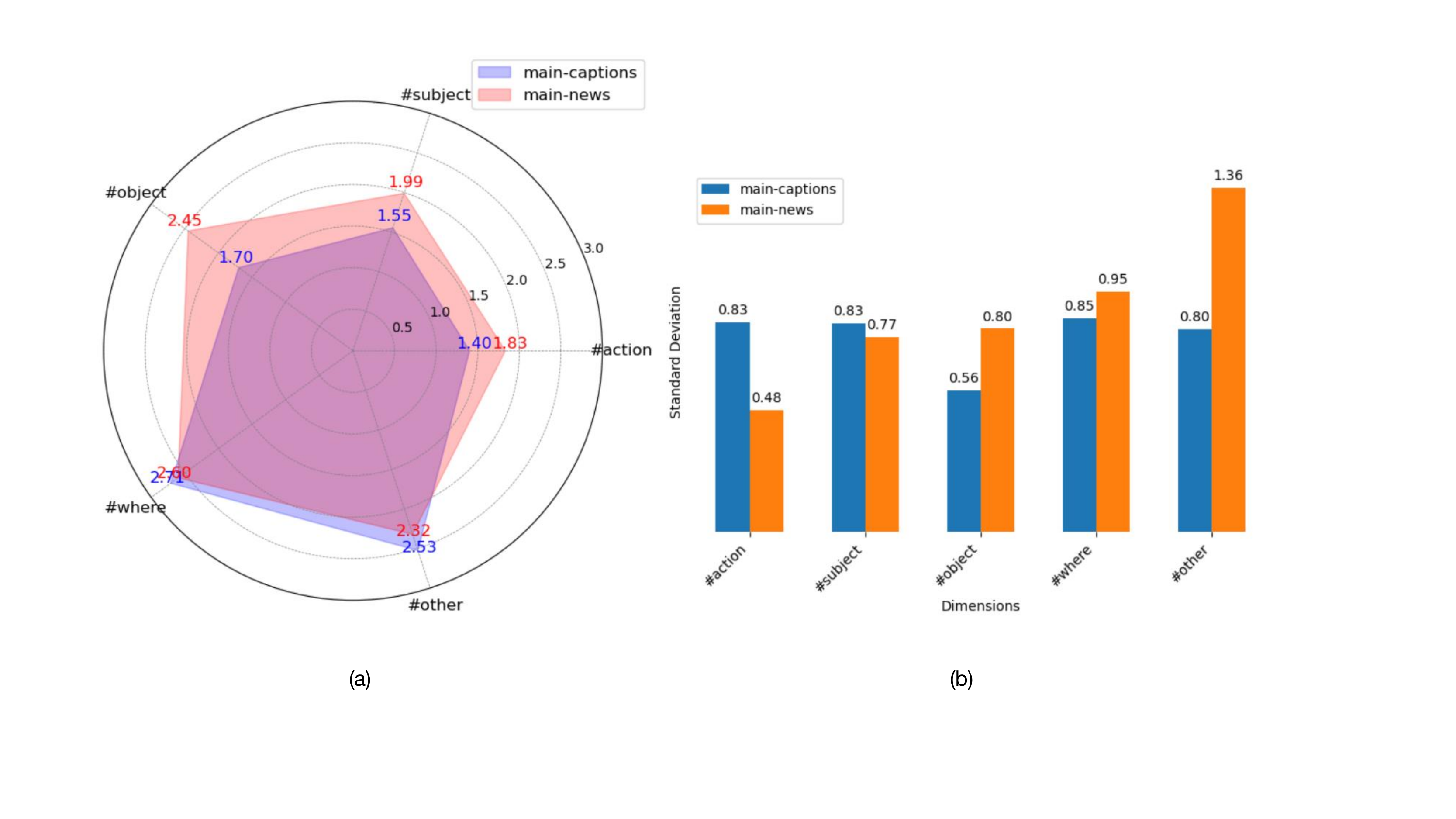}
  \caption{(a) Mean Ground Scores. For \texttt{main-captions}, disparities in \texttt{\#action}, \texttt{\#subject} or \texttt{\#object} tend to make the two sentences appear more dissimilar than \texttt{\#where} and \texttt{\#other}. Additionally, a difference in \texttt{\#action}, \texttt{\#subject} or \texttt{\#object} is a sufficient condition for the non-equivalence of two propositions in human perception of sentence similarity. (b) Standard Deviation.}
  \label{fig:cap_vs_news}
\end{figure*}



This section studies the cognitive process underlying how humans assign a score according to task instructions when given a sentence pair. Intuitively, different dimensions ought to contribute differently to the disparity between two sentences. In addition, for the same dimension, human perceptions of the disparity should vary to a certain extent. We aim to quantify these cognitive differences in this experiment. 

Eight dimensions from Section~\ref{pair_type_p1} are considered: \texttt{\#action}, \texttt{\#subject}, \texttt{\#object}, \texttt{\#where}, \texttt{\#aux\_obj}, \texttt{\#goal}, \texttt{\#source}, and \texttt{\#reason}. We aggregate \texttt{\#aux\_obj}, \texttt{\#goal}, \texttt{\#source}, and \texttt{\#reason} into one dimension, labeled as \texttt{\#other}. 

\textbf{Dataset.} The \texttt{P1-} pairs of the STS-B dataset, whose sentences differ at only one dimension, are selected. Compared to the black-box nature of embedding models, PropNet significantly facilitates this selection process by simply checking if the corresponding element of difference vector is 2. For example, "A man is speaking" and "A man is cooking" is a valid pair for the \texttt{\#action} dimension, since the difference vector has a value 2 at \texttt{\#action} and 0 at all other positions. Only pairs from \texttt{main-captions} and \texttt{main-news} are used, since \texttt{main-forums} contains too few samples. In total, 133 valid pairs are collected for analysis. All pairs and their distribution are detailed in Appendix~\ref{app:experiment_cog_diff_samples}.

\textbf{Evaluation Metrics.} The mean values of the ground-truth scores of the selected pairs within each dimension are calculated to reflect the human cognitive weight assigned to sentence similarity across the five dimensions. Additionally, the standard deviations of each dimension are provided to measure the cognitive deviations within a single dimension.

\textbf{Results.} In Figure~\ref{fig:cap_vs_news} (a), for \texttt{main-captions}, the mean values of \texttt{\#action}, \texttt{\#subject} and \texttt{\#object} are around 1.5, which are lower than those of \texttt{\#where} and \texttt{\#other} (around 2.5), with statistical significance at the 0.05 level. This indicates that disparities in \texttt{\#action}, \texttt{\#subject} or \texttt{\#object} make the two sentences seem more different than the disparities in other two dimensions. For \texttt{main-news}, all dimensions have similar weights on difference perception, with no statistical significance. 

Another discovery is that pairs with disparity within the dimension \texttt{\#action}, \texttt{\#subject}, or \texttt{\#object} in both genres have ground scores below 2.5, which is the cut-off score for ``not equivalent'' and ``roughly equivalent''. This implies that a difference in any of these three dimensions is a sufficient condition for the non-equivalence of two propositions in human perception of caption or news sentence similarity. The ground score rules are provided in Appendix~\ref{app:experiment_cog_diff_rules}. Research regarding more complex sentence pairs, such as \texttt{P2}, is deferred to future work.

Figure~\ref{fig:cap_vs_news} (b) illustrates the differences in standard deviation for manual scoring. No statistically significant human cognitive deviation is observed within any single dimension at the 0.05 level. Overall, \texttt{main-captions} exhibits a standard deviation of approximately 0.8. For \texttt{main-news}, the notable difference between \texttt{\#action} and \texttt{\#other} presented in the figure is only statistically significant at the 0.1 level. Future work aims to collect more samples to reevaluate cognitive deviations within individual dimensions. All hypothesis testing results are supplied in Appendix~\ref{app:experiment_cog_diff_testing}.

\section{Discussion}

\textbf{Multimodal Foundation.} Instance nodes are necessary not only to process language, but also to empower PropNet with the potential to function as the backbone of understanding human's vision and control. Imagine this: you see an apple on the table and stretch out your arm to pick it up. An instance node corresponding to this apple makes it possible to aggregate shape, color and name information together, distinguish this apple from other apples on the table, and also possible to track this apple and compute its relative positions to other objects when it is picked up. 

\textbf{Text Generation.} If PropNet can represent an image or a video, the rules used in the splitting and parsing phases can be reversed to generate a meaningful sentence. For instance, given the subject "apple", the action "fall", and the action source "tree" perceived from an image or a video, PropNet can generate the sentence "apple fall from tree". This represents a novel approach to generate text, fundamentally distinct from the current mainstream multimodal large language models. It is completely transparent and is likely to be closer to the way humans generate language.

\textbf{Long Text Understanding.}  Although in this work we mainly discuss the representation of a single sentence, it is possible to extend this method to a discourse or a conversation. Instance nodes enable PropNet to have potentials in addressing coreference resolution and representing contextual information effectively. Minor changes to current merging strategies can meet the need for linking the representations of each sentence into a unified network. This will provide novel perspectives for the research of long-text understanding in a white-box and human-like manner.

\textbf{Cognitive Process Analysis.} With the emergence of deep learning algorithms, the past decade has witnessed the generation of massive amounts of manually labeled data for training models. Therefore, cognitive processes underlying how humans label input data according to specific instructions, are recorded in these vast datasets. White-box and human-like approaches have the potential to illuminate the cognitive processes, which is exemplified by Section~\ref{cog_diff_in_sts}. It can significantly contribute to a deeper understanding of humanity, such as our behaviors, mentality, and intelligence.

\section{Limitation}

PropNet would fail to represent if there are errors from spaCy part-of-speech tagging and dependency parsing results. For instance, in the sentence ``Blizzard wallops US Northeast, closes road'', the subject of ``close'' is identified as ``wallop''.

\section{Conclusion}
In this paper, we have proposed a novel sentence representation network called PropNet. It can represent a complex sentence by a hierarchical network with interpretable and transparent structure. Through case analysis, several feasible optimization directions are identified to narrow the gap with SOTA models in STS tasks. PropNet paves the way for analyzing and understanding the human cognitive processes underlying STS benchmarks.

\bibliographystyle{plain}
\bibliography{custom}

\clearpage
\appendix
\section{PropNet Construction}
\subsection{Cases}
\label{app:propnet_cases}

\begin{figure*}[h!]
  \centering
  \includegraphics[width=0.8\textwidth]{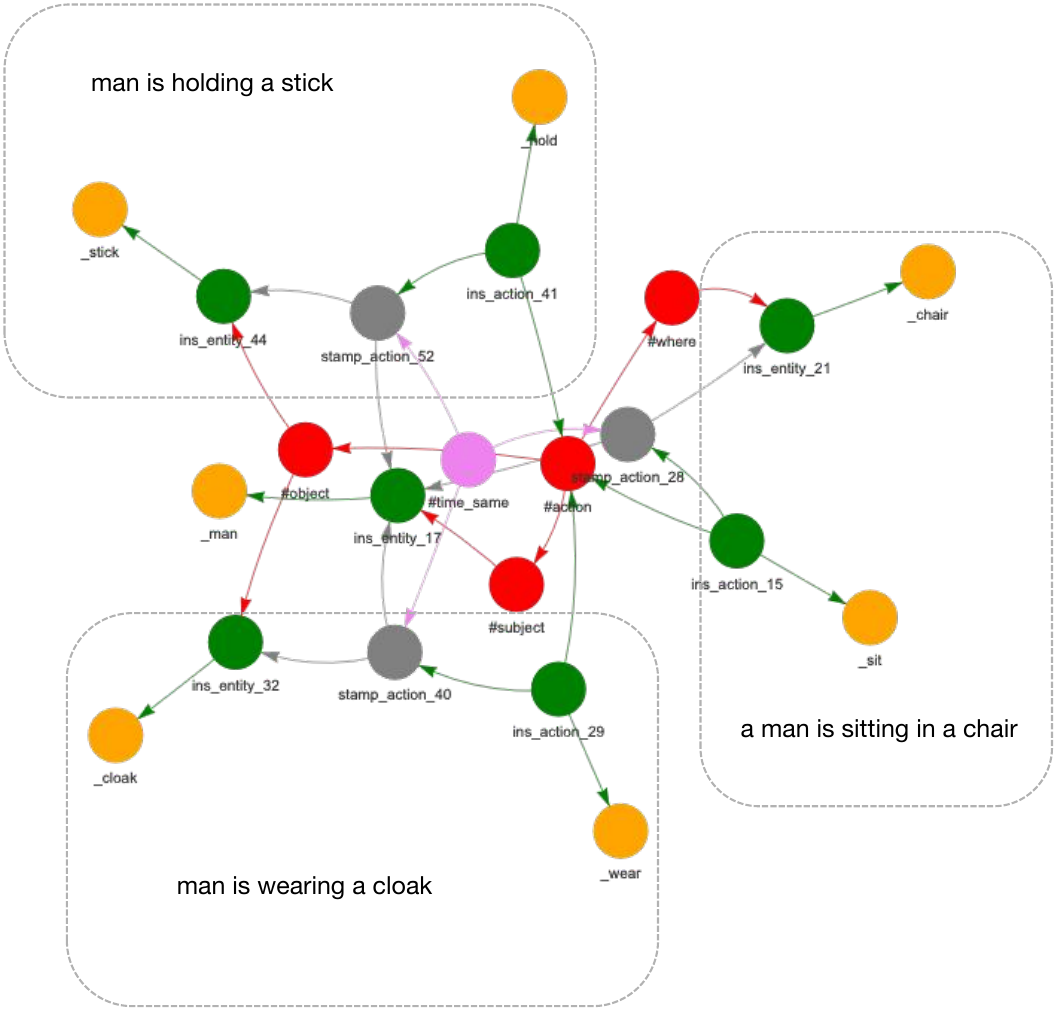}
  \caption{PropNet for ``A man is sitting in a chair wearing a cloak and holding a stick.'' It merges the representations of three propositions: ``a man is sitting in a chair'', ``man be wearing a cloak'', and ``man be holding a stick'', each enclosed by separate grey frames.}
  \label{fig:cs_3v_repr}
\end{figure*}

Figure~\ref{fig:cs_3v_repr} shows the PropNet representation of a sentence with three propositions. The instance nodes pointing to \texttt{\_unknown} are all omitted, as well as their evolutionary nodes, in order to provide a clean view. From this representation, we can form a picture of what happens in our mind.

\begin{figure*}[h!]
  \includegraphics[width=1\textwidth]{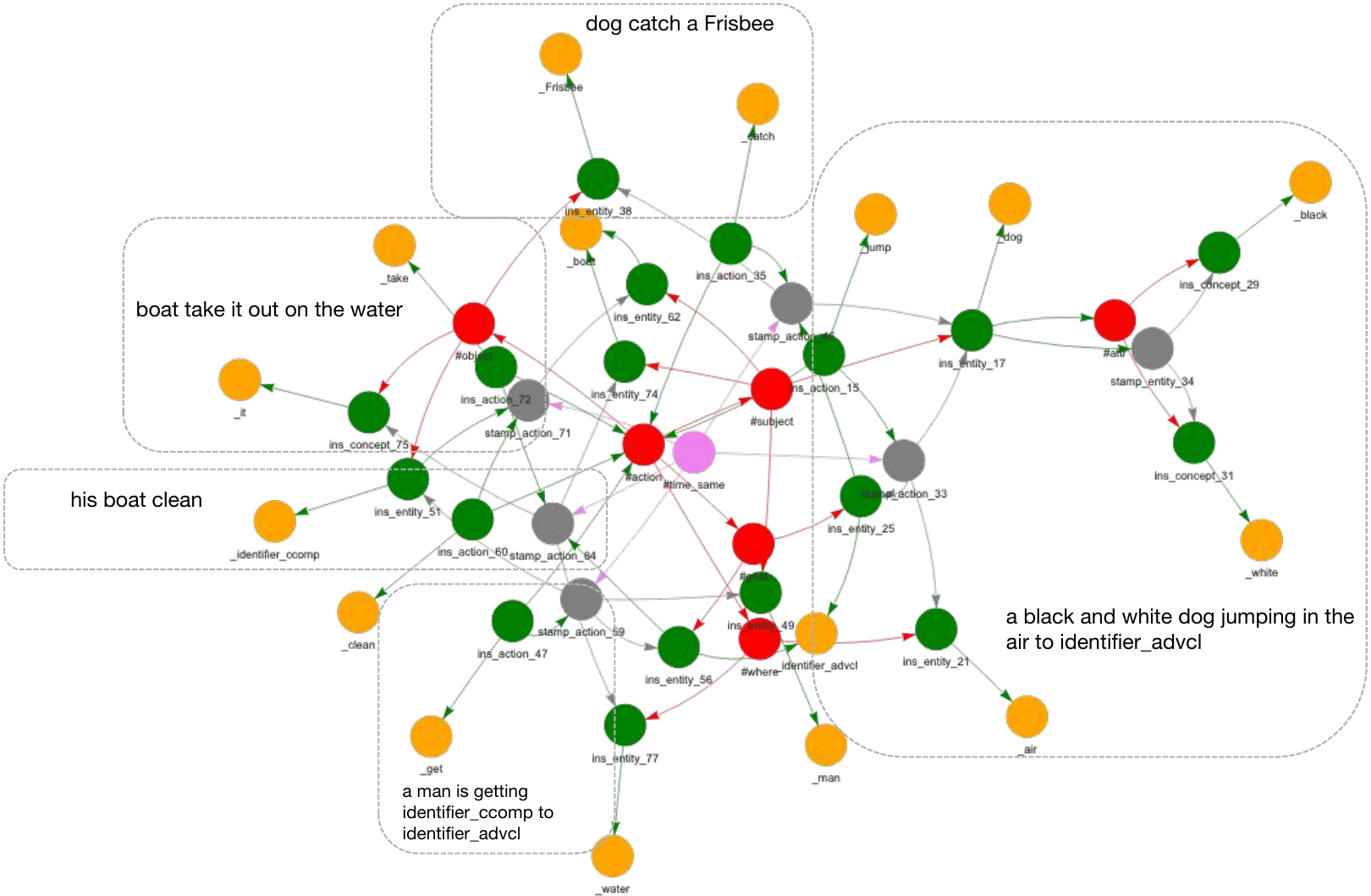}
  \caption{PropNet for ``A black and white dog jumping in the air to catch a Frisbee when a man is getting his boat clean to take it out on the water.'' It merges the representations of five propositions: ``a black and white dog jumping in the air to identifier\_advcl'', ``dog catch a Frisbee'', ``a man is getting identifier\_ccomp to identifier\_advcl'', ``his boat clean'', and ``boat take it out on the water'', each enclosed within five grey frames.}
  \label{fig:cs_5v_repr}
\end{figure*}

A more complex sentence with five propositions is presented in Figure~\ref{fig:cs_5v_repr}. Its splitting and merging processes are explained in Figure~\ref{fig:split_and_merge_cs_plus3v}. Notably, the subject of the last proposition is incorrect. The correct result should be ``man takes it out on the water''. A more robust strategy of subject identification in the splitting phase will be added in future work.

\subsection{Relative Clause}
\label{app:propnet_relative_clause}

\begin{figure*}[h!]
  \centering
  \includegraphics[width=0.8\textwidth]{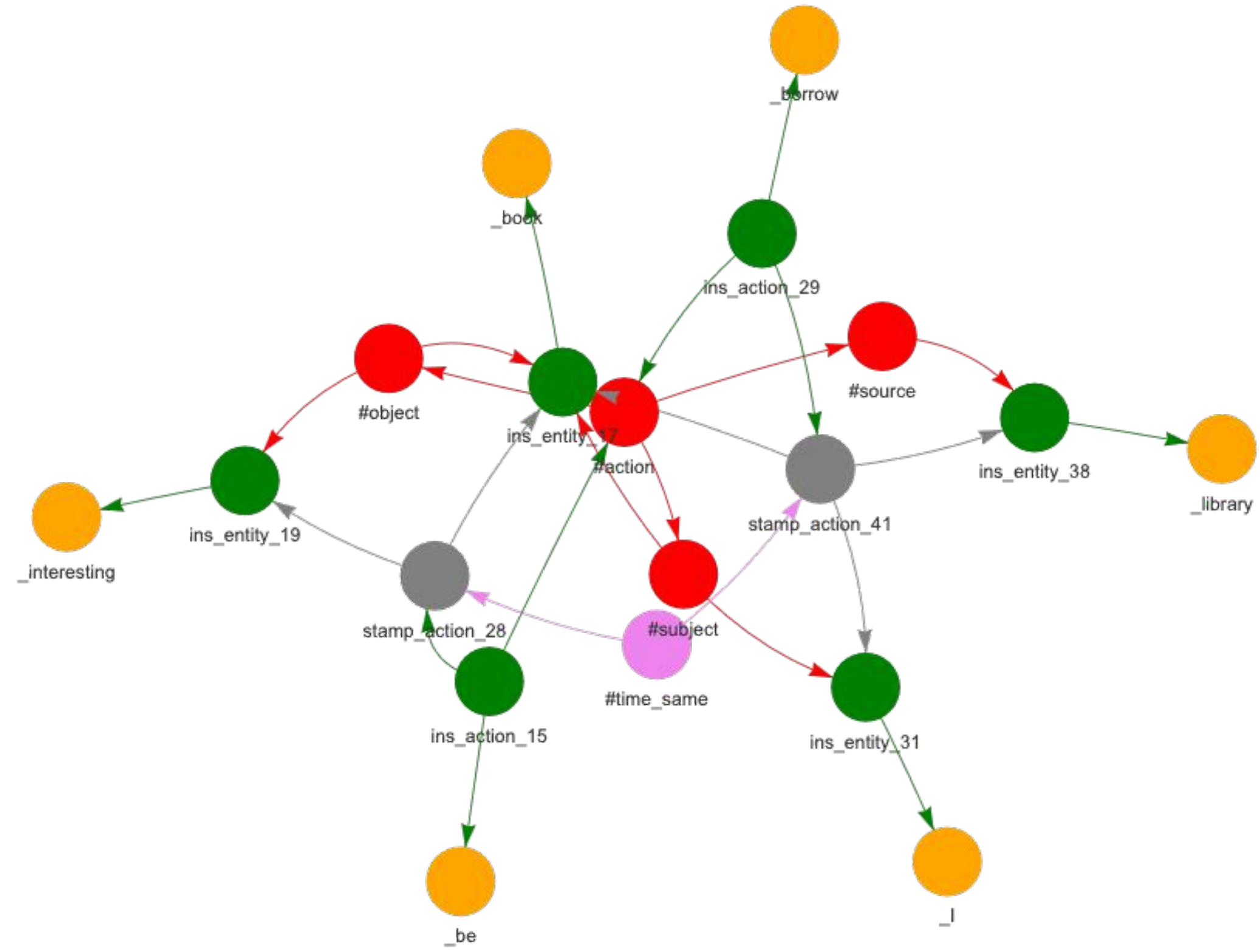}
  \caption{A relative pronoun example ``The book that I borrowed from the library is interesting.'' It is split into two propositions: ``the book is interesting'', ``I borrowed book from the library''. The instance nodes of \textbf{book} are merged into \texttt{ins\_entity\_17}.}
  \label{fig:relative_pronoun}
\end{figure*}

\begin{figure*}[h!]
  \centering
  \includegraphics[width=0.8\textwidth]{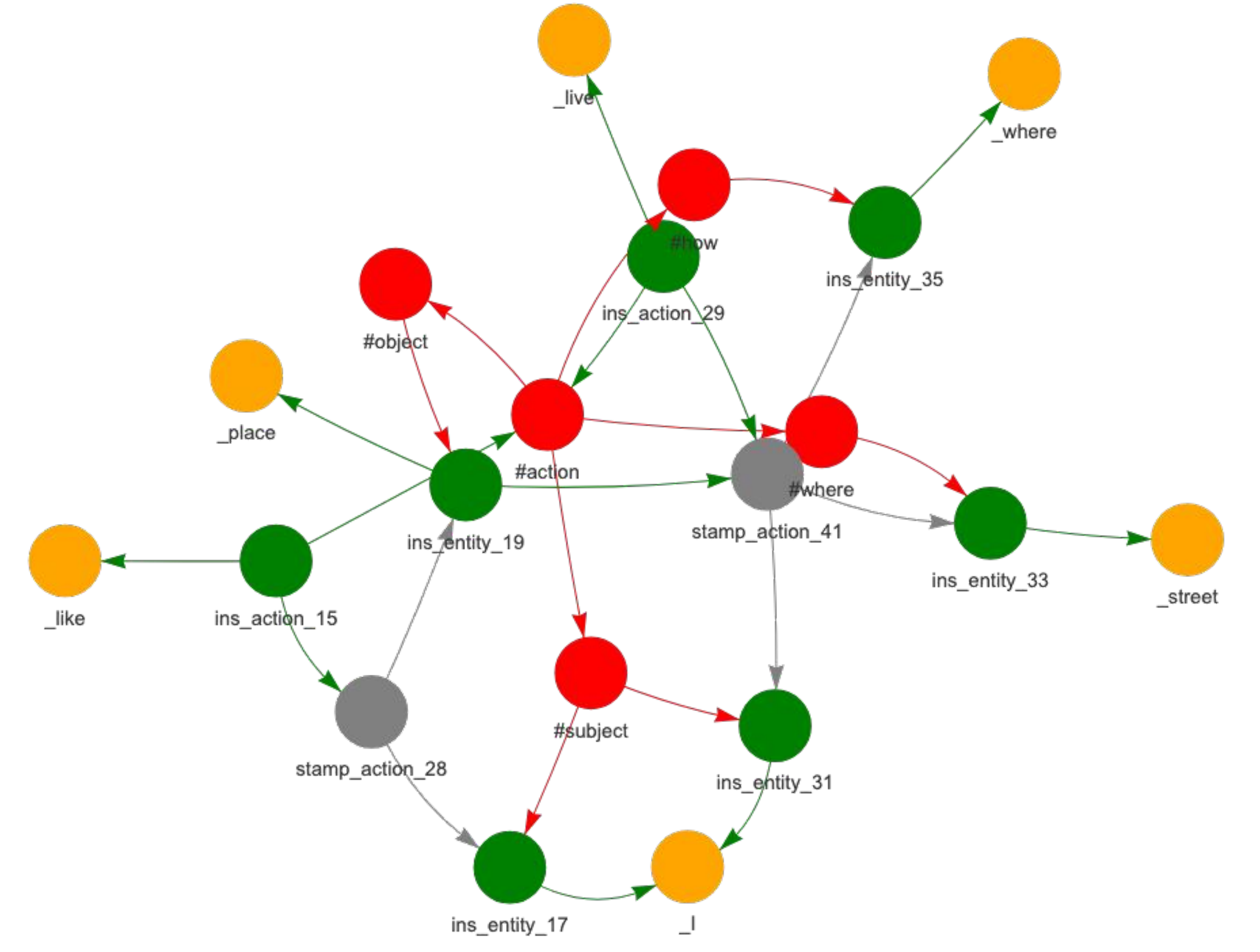}
  \caption{A relative adverb example ``I like the place where I live in the street.'' It is split into two propositions: ``I like the place'', ``where I live in the street''. The instance node of \textbf{place}, \texttt{ins\_entity\_19}, is linked to the action stamp node \texttt{stamp\_action\_41}, invoked by \textbf{live}.}
  \label{fig:relative_adv}
\end{figure*}

\begin{figure*}[h!]
  \centering
  \includegraphics[width=0.8\textwidth]{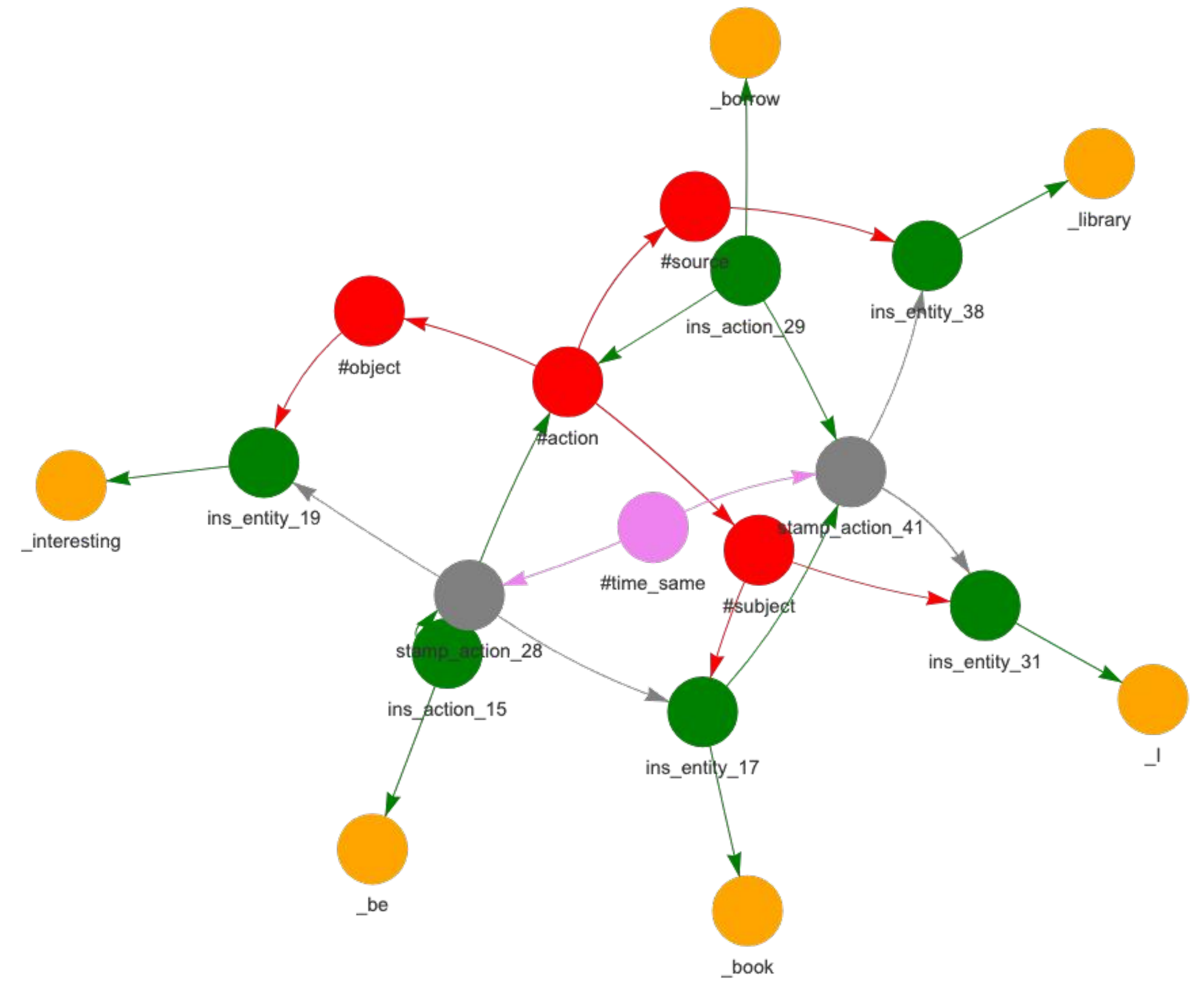}
  \caption{A no-relative example ``The book I borrowed from the library is interesting.'' It is split into two propositions: ``the book is interesting'', ``I borrowed from the library''. The instance node of \textbf{book}, \texttt{ins\_entity\_17}, is linked to the action stamp node \texttt{stamp\_action\_41}, invoked by \textbf{borrow}.}
  \label{fig:no_relative}
\end{figure*}

As a relative clause is more complex than other types of clauses, additional rules are proposed to address its splitting and merging. Currently, only relative pronouns \texttt{that}, \texttt{which}, \texttt{who} and relative adverbs \texttt{when}, \texttt{where} are considered for a \texttt{Prop2} sentence with a relative clause. Table~\ref{tab:splitting_rules_cs2v_relcl} shows the splitting and merging rules with examples. The merging results of these examples are shown in Figure~\ref{fig:relative_pronoun}, Figure~\ref{fig:relative_adv} and Figure~\ref{fig:no_relative}, respectively. It should be noted that the action stamp nodes of main proposition and subordinate proposition are linked by the evolutionary node \texttt{\#time\_same}. This merging rule is not included in the table.

\begin{table*}[h]
\centering
\caption{Splitting and Merging Rules for a \texttt{Prop2} Sentence with a Relative Clause.}
\label{tab:splitting_rules_cs2v_relcl}
\small 
\renewcommand{\arraystretch}{1.5}
\begin{tabular}{@{}p{1.2cm}p{4.5cm}p{2.7cm}p{4.8cm}@{}}
\toprule
\textbf{Type} & \textbf{Splitting Rule} & \textbf{Merging Rule} & \textbf{Example} \\ \midrule

Relative Pronouns & If \texttt{t.lemma\_ in (\textbf{that}, \textbf{which}, \textbf{who})} and \texttt{t.dep\_ in (\textbf{nsubj}, \textbf{dobj})}, replace \texttt{t} by the antecedent in subordinate proposition. & Merge the two instance nodes of antecedent. & The \textbf{book} that I borrowed from the library is interesting. -> (the \textbf{book} is interesting, I borrowed \textbf{book} from the library) [Merge instance nodes of \textbf{book}.]\\

Relative Adverbs & If \texttt{t.lemma\_ in (\textbf{when}, \textbf{where})} and \texttt{t.dep\_ in (\textbf{advmod})}, keep \texttt{t} in subordinate proposition. & Link instance node of antecedent to action stamp node of subordinate proposition. & I like the \textbf{place} where I live in the street. -> (I like the \textbf{place}, where I live in the street) [Link instance node of \textbf{place} to action stamp node invoked by \textbf{live}]\\
No Relatives & No processing in subordinate proposition. & Link instance node of antecedent to action stamp node of subordinate proposition. & The \textbf{book} I borrowed from the library is interesting. -> (the \textbf{book} is interesting, I borrowed from the library) [Link instance node of \textbf{book} to action stamp node invoked by \textbf{borrowed}]\\ \bottomrule
\end{tabular}
\end{table*}

\subsection{Parsing}
\label{app:propnet_parse}

This section provides a detailed explanation of the primary parsing rules. For  
\texttt{Action}, \texttt{Subject}, \texttt{Object}, \texttt{Attribute} and \texttt{Part\_of}, the parsing primarily relies on \texttt{token.dep\_} and \texttt{token.pos\_}. For \texttt{Where}, \texttt{Auxiliary\_Object}, \texttt{Goal}, \texttt{Reason} and \texttt{Source}, the parsing system depends on detecting and classifying token's prepositions. The classification is based on the definitions of prepositions from the Merriam-Webster Dictionary. We acknowledge that current classification remains course and occasionally inaccurate. For example, in ``shoot at an elephant'', ``elephant'' is classified as \texttt{Where}. However, this \texttt{Where} should be interpreted as the action goal in conjunction with the meaning of "shoot". This issue can be addressed by adding more evolutionary nodes. Additionally, the temporal attributes of an action, such as its timing, are temporarily assigned to \texttt{Where}. More sophisticated space-time modules are planned for development in future research. 

The complete list of preposition categories is as follows:

\begin{itemize}
    \item \texttt{Where}: ``on'', ``in'', ``inside'', ``through'', ``over'', ``around'', ``down'', ``at'', ``near'', ``along'', ``outside'', ``past'', ``across'', ``during'', ``after'', ``before'', ``while'', ``whilst'', ``off'', ``amid'', ``behind''. 
    \item \texttt{Auxiliary\_Object}: ``with'', ``by'', ``about'', ``like'', ``as''.
    \item \texttt{Goal}: ``into'', ``to'', ``onto'', ``towards'', ``against''
    \item \texttt{Reason}: ``for'', ``due''.
    \item \texttt{Source}: ``from''.
    
\end{itemize}

\section{Comparing Two Sentences}
\subsection{Comparison Dimensions}
\label{sec:compare_dim}

Table~\ref{tab:dimensions_diff_vec} explains the comparison dimensions to compute a difference vector given two sentence representations. Be aware that since instance nodes themselves have no meanings, what are really compared are the corresponding developmental nodes.

\begin{table*}[h]
\centering
\begin{tabular}{@{}l@{}p{10cm}@{}}
\toprule
\textbf{Dimension} & \textbf{Explanation} \\
\midrule
\texttt{\#action} & Find and compare all the instance nodes that represent actions. \\
\midrule
\texttt{\#action|\#subject} & Find and compare all the instance nodes that represent the subject of an action. \\
\texttt{\#action|\#subject|\#attr} & Find and compare all the instance nodes that belong to attributes of an entity or concept that serves as the subject of the action. \\
\texttt{\#action|\#subject|\#part\_of} & Find and compare all the instance nodes that represent parts of an entity or concept that serves as the subject of the action. \\
\texttt{\#action|\#subject|\#where} & Find and compare all the instance nodes that represent the location of the subject of the action. \\
\midrule

\texttt{\#action|\#object} & Find and compare all the instance nodes that represent the object of an action. \\
\texttt{\#action|\#object|\#attr} & Find and compare all the instance nodes that belong to attributes of an entity or concept that serves as the object of the action. \\
\texttt{\#action|\#object|\#part\_of} & Find and compare all the instance nodes that represent parts of an entity or concept that serves as the object of the action. \\
\texttt{\#action|\#object|\#where} & Find and compare all the instance nodes that represent the location of the object of the action. \\
\midrule

\texttt{\#action|\#aux\_obj} & Find and compare all the instance nodes that represent auxiliary objects of an action. \\
\texttt{\#action|\#aux\_obj|\#attr} & Find and compare all the instance nodes that belong to attributes of an entity or concept that serves as the auxiliary object of the action. \\
\texttt{\#action|\#aux\_obj|\#part\_of} & Find and compare all the instance nodes that represent parts of an entity or concept that serves as the auxiliary object of the action. \\
\midrule

\texttt{\#action|\#where} & Find and compare all the instance nodes that represent the location of the action. \\
\texttt{\#action|\#where|\#attr} & Find and compare all the instance nodes that belong to attributes of the location of the action. \\
\texttt{\#action|\#where|\#part\_of} & Find and compare all the instance nodes that represent parts of the location of the action. \\
\midrule

\texttt{\#action|\#goal} & Find and compare all the instance nodes that represent the goal of the action. \\
\texttt{\#action|\#goal|\#attr} & Find and compare all the instance nodes that belong to attributes of the goal of the action. \\
\texttt{\#action|\#goal|\#part\_of} & Find and compare all the instance nodes that represent parts of the goal of the action. \\
\midrule

\texttt{\#action|\#source} & Find and compare all the instance nodes that represent the source of the action. \\
\texttt{\#action|\#source|\#attr} & Find and compare all the instance nodes that belong to attributes of the source of the action. \\
\texttt{\#action|\#source|\#part\_of} & Find and compare all the instance nodes that represent parts of the source of the action. \\
\midrule

\texttt{\#action|\#reason} & Find and compare all the instance nodes that represent the reason for the action. \\
\texttt{\#action|\#reason|\#attr} & Find and compare all the instance nodes that belong to attributes of the reason for the action. \\
\texttt{\#action|\#reason|\#part\_of} & Find and compare all the instance nodes that represent parts of the reason for the action. \\
\bottomrule
\end{tabular}
\caption{Comparison Dimensions for Computing A Difference Vector. }
\label{tab:dimensions_diff_vec}
\end{table*}

\subsection{Similarity between Two Words}
\label{app:llm_lexical_similarity}

Given two words, we use Doubao \footnote{\url{https://team.doubao.com/en/direction/llm}} to compute a similarity score ranging from 0 to 1. Doubao has an extremely low inference cost, while offering top-tier inference performance and an easy-to-use API for calling. The quality of the output scores are validated by manual inspection during development. Since this task is straightforward, other LLM models like ChatGPT or Llama should work as well. 

For a verb pair, its prompt is "From physical action perspective, return similarity score (0\textasciitilde1) between two verbs: \{word\_text1\} and \{word\_text2\}. Only return the score".

For other pairs, its prompt is "From semantic perspective, return similarity score (0\textasciitilde1) between two words: \{word\_text1\} and \{word\_text2\}. Only return the score".

\begin{table*}[h!]
\centering
\renewcommand{\arraystretch}{2}
\small 
\begin{tabular}{|p{0.04\textwidth}|p{0.13\textwidth}|p{0.35\textwidth}|p{0.3\textwidth}|}
\hline
\textbf{Code} & \textbf{Meaning} &\textbf{Calculation Rules} & \textbf{Examples} \\
\hline
0 & Completely or Almost the Same & 
\makecell[l]{
(1) Two words have the same text; \\
(2) LLM similarity score >= 0.7; \\
(3) binary group similarity score >= 0.7
} 
& 
\makecell[l]{
(1) "guitar" vs "guitar"; \\
(2) "cat" vs "kitten" \\
(3) ["red", "cat"] vs ["pink", "cat"]
} \\
\hline
1  & Similar & 
\makecell[l]{
(1) 0.2 <= LLM similarity score < 0.7; \\
(2) One side is "unknown"; \\
(3) 0.2 <= binary group similarity score < 0.7\\
}
& 
\makecell[l]{
(1) "guitar" vs "piano"; \\
(2) "guitar" vs "unknown"; \\
"unknown" vs ["guitar", "piano"] \\
(3) ["short", "man"] vs ["tall", "man"]
} \\
\hline

2 & Different & \makecell[l]{(1) LLM similarity score < 0.2 \\ (2) binary group similarity score < 0.2} & \makecell[l]{(1) "tall" vs "short" \\ (2) ["short", "man"] vs ["big", "cat"]} \\
\hline
3 & Long and Equal Length & \makecell[l]{Two group words with equal length \\ which is greater than 2.} & \makecell[l]{["violin", "guitar", "piano"] \\ vs ["car", "bus", "truck"]} \\
\hline
4 & Different Length & \makecell[l]{(1) Single word and a group of words; \\(2) Two group words with different length.} & \makecell[l]{(1) "guitar" vs ["violin", "piano"] \\ (2) ["guitar", "piano"] \\ vs ["car", "bus", "truck"]} \\
\hline

\end{tabular}
\caption{Dimensional Similarity Code. }
\label{tab:similar_code}
\end{table*}

\subsection{Dimensional Similarity Code}
\label{app:similar_code}

Given a comparison dimension, it's common that both sentences have only one developmental node. For example, considering this pair "I like orange" and "she hates apple", \texttt{\#action} is "like" and "hates", \texttt{\#action|\#subject} is "I" and "she", and \texttt{\#action|\#object} is "orange" and "apple". But exceptions do exist that the developmental nodes might be more than one. For instance, in the representation of "I like orange and apple", \texttt{\#action|\#object} has two developmental nodes \texttt{\_orange} and \texttt{\_apple}. 

We define Dimensional Similarity Code (DSC) as an indicator for assessing similarity between two sentence representations under a specific comparison dimension. Three scenarios are considered: (1) comparing two words, (2) comparing a single word with a group of words, and (3) comparing two groups of words. The DSC codes and their corresponding computing rules are explained in Table~\ref{tab:similar_code}. 

Given two words, a code of 0 means two words are completely or almost the same. A code of 1 means they are similar, sharing some important details. Note that when we compare "unknown" with one single word or a group of words, they are regarded as similar. For example, in the comparison of "I cut an onion with a knife" and "I cut an onion", the dimension \texttt{\#action|\#aux\_obj} has "knife" for the first sentence and "unknown" for the second sentence, and this comparison receives a code of 1. A code of 2 means the two words are completely different. We manually check word pair cases along with their LLM scores to determine proper thresholds, best matching the DSC code standard. If LLM similarity score is no less than 0.7, a code of 0 is assigned. If LLM similarity score is less than 0.2, a code of 2 is assigned. Otherwise, a code of 1 is assigned.


When comparing one single word with a group of words, or two groups of words of different lengths, we label this scenario as "different length" with a code of 4. An exception occurs when the single word is "unknown", in which case it receives a code of 1. Further refinement of similarity discrimination, incorporating semantic analysis, is left for future work.

Given two groups of words with equal length greater than 2, we label this situation as "long and equal length" and assign it a code of 3. More refined similarity discrimination, combined with semantics, is deferred to future work.

Finally, given two groups of words with an equal length 2, we consider a binary group similarity score which is calculated by the following steps:

\begin{enumerate}[label=(\arabic*)]
    \item Initialize \texttt{score} as 0.
    \item Directly calculate the intersection between two groups of words and obtain the length of the intersection.
    \item Add the length of the intersection to the \texttt{score} variable.
    \item Remove the intersecting parts from each word group.
    \item If each group contains only one remaining word, calculate the similarity score between these two words and add it to \texttt{score}. Otherwise do nothing.
    \item Divide \texttt{score} by 2 to obtain the binary group similarity score.
\end{enumerate}

This process is exemplified step by step by comparing ["apple", "banana"] and ["apple", "orange"] as follows:

\begin{enumerate}[label=(\arabic*)]
    \item Initialize \texttt{score} as 0.
    \item Intersection is ["apple"] and its length is 1.
    \item  \texttt{score} = 1 + \texttt{score}. Now \texttt{score} is 1.
    \item Remove "apple" from both groups. Now they become ["banana"] and ["orange"].
    \item Since each group has only one word left, LLM is used to calculate the similarity score which outputs a 0.2. Now \texttt{score} is 1.2.
    \item Binary group similarity score = score / 2 = 0.6
\end{enumerate}

As the final score lies between 0.2 and 0.7, a DSC code 1 (similar) is assigned for this binary-word groups comparison.

\section{Experiments}
\subsection{CART Training Details for STS Tasks}
\label{app:cart_train_details}

For the STS-B dataset, we tune only one parameter for CART training, the minimum number of samples at a leaf node (short as \texttt{min\_samples\_leaf}), on its \texttt{train.jsonl} (5749 samples) and \texttt{validation.jsonl} (1500 samples). The default values are adopted for other CART parameters, since finetuning models for higher Spearman's score is not a key point in our work. What we place high weight on, is finding the weaknesses of PropNet and its opitimization directions in future research. 

The SICK-R dataset contains a field named \texttt{SemEval\_set}, indicating the train/dev/test split. We use the training split (4439 samples) and dev split (495 samples) for tuning \texttt{min\_samples\_leaf}. 

CART Training is supported by the Scikit-Learn APIs, whose version is 1.6.0.

\subsection{Genres}
\label{app:genres}
STS-B dataset has a field \texttt{genre} with three values, which are explained as follows:

\begin{itemize}
    \item \texttt{main-captions}: image/video description, focusing on the main action or event in a video or an image. For example, "A woman is spreading mustard on a bread roll." The words have strong relations with physical concepts, as defined in \citep{10692615}.
    
    \item \texttt{main-news}: news, for instance, "Stones add second Hyde Park concert". The words have strong relations with social concepts.
    
    \item \texttt{main-forums}: forum question or answer, for instance, "This is a part answer to your question". The words appearing in such a sentence are mainly \texttt{mental}, as introduced in \citep{10692615}. 
\end{itemize}

Based on the action types introduced in \citep{Yang2024AutomaticEO}, a proposition of \texttt{main-captions}, \texttt{main-news} and \texttt{main-forums} correspond to \texttt{physical}, \texttt{mental} and \texttt{social} types, respectively.


\subsection{Cognitive Differences in STS Tasks}
\label{app:experiment_cog_diff}
\subsubsection{Experimental Samples}
\label{app:experiment_cog_diff_samples}

Due to insufficient pairs for \texttt{\#aux\_obj}, \texttt{\#goal}, \texttt{\#source}, and \texttt{\#reason}, these four dimensions are aggregated into one, labeled as \texttt{\#other}. Valid pairs from STS 2012-2016 are added for the \texttt{\#other} dimension. The experiment’s results are not affected by this supplement, since the scoring criteria remain consistent across STS 2012-2016 and STS-B datasets. 

All pair examples selected for \texttt{main-captions} and \texttt{main-news} in this experiment are listed in Table~\ref{tab:exp3_caption_examples} and Table~\ref{tab:exp3_news_examples} respectively. Pair distributions,  ground score means and standard deviations for \texttt{main-captions} and \texttt{main-news} are listed in Table~\ref{tab:exp3-stat-main-captions} and Table~\ref{tab:exp3-stat-main-news} respectively. 

\begin{table}[H]
\centering
\begin{tabular}{@{}p{1.5cm}p{1cm}p{1.5cm}p{0.8cm}p{0.8cm}@{}}
\toprule
\textbf{Dimension} & \textbf{Count} & \textbf{Proportion (\%)} & \textbf{Mean} & \textbf{STD} \\
\midrule

\texttt{\#action} & 24 & 27.91 & 1.40 & 0.83 \\
\texttt{\#subject} & 12 & 13.95 & 1.55 & 0.83 \\
\texttt{\#object} & 25 & 29.07 & 1.70 & 0.56 \\
\texttt{\#where} & 21 & 24.42 & 2.71 & 0.85 \\
\texttt{\#other} & 4 & 4.65 & 2.53 & 0.80 \\

\bottomrule
\end{tabular}
\caption{Pair statistics for \texttt{main-captions}. The total number of pairs is 86.}
\label{tab:exp3-stat-main-captions}
\end{table}

\begin{table}[H]
\centering
\begin{tabular}{@{}p{1.5cm}p{1cm}p{1.5cm}p{0.8cm}p{0.8cm}@{}}
\toprule
\textbf{Dimension} & \textbf{Count} & \textbf{Proportion (\%)} & \textbf{Mean} & \textbf{STD} \\
\midrule

\texttt{\#action}           & 6              & 12.77                    & 1.83          & 0.48         \\ \hline
\texttt{\#subject} & 10             & 21.28                    & 1.99          & 0.77         \\ \hline
\texttt{\#object}  & 12             & 25.53                    & 2.45          & 0.80         \\ \hline
\texttt{\#where}   & 9              & 19.15                    & 2.60          & 0.95         \\ \hline
\texttt{\#other}             & 10             & 21.28                    & 2.32          & 1.36         \\ 
\bottomrule
\end{tabular}
\caption{Pair statistics for \texttt{main-news}. The total number of pairs is 47.}
\label{tab:exp3-stat-main-news}
\end{table}

\subsubsection{Statistical Testing}
\label{app:experiment_cog_diff_testing}
Mann-Whitney U Test is performed to detect the statistical difference in ground score means between any two dimensions, as the ground scores are not normally distributed for all dimensions using Shapiro-Wilk Test. The significance level is set to 0.05. Results for \texttt{main-captions} are reported in Table~\ref{tab:u_test_caption}, indicating \texttt{\#action}, \texttt{\#subject} and \texttt{\#object} are significantly different from \texttt{\#where} and \texttt{\#other}. Although \texttt{\#subject} and \texttt{\#other} do not reach the significance level of 0.05, they are close to the significance level of 0.1. The situation for \texttt{\#object} and \texttt{\#other} is also similar. Results for \texttt{main-news} are reported in Table~\ref{tab:u_test_news}, revealing that there is no statistical significance.

The Levene's Test is conducted to assess the statistical difference in the ground score standard deviations between all pairs of dimensions. The significance level is set to 0.05. The results for \texttt{main-captions} and \texttt{main-news} are reported in Table~\ref{tab:levene_test_caption} and Table~\ref{tab:levene_test_news}, respectively. We do not find any statistically significant differences. 

\begin{table*}[h]
\centering
\begin{tabular}{|c|c|c|c|c|}
\hline
\textbf{Dimension1} & \textbf{Dimension2} & \textbf{U-statistic} & \textbf{P-value} & \textbf{Statistical Significance (0.05)} \\
\hline
\#action & \#subject & 121.5 & 0.4595 & No \\
\hline
\#action & \#object & 217.5 & 0.1000 & No \\
\hline
\textbf{\#action} & \textbf{\#where} & \textbf{73.0} & \textbf{0.0000} & \textbf{Yes} \\
\hline
\textbf{\#action} & \textbf{\#other} & \textbf{16.5} & \textbf{0.0413} & \textbf{Yes} \\
\hline
\#subject & \#object & 126.5 & 0.4545 & No \\
\hline
\textbf{\#subject} & \textbf{\#where} & \textbf{47.0} & \textbf{0.0032} & \textbf{Yes} \\
\hline
\#subject & \#other & 10.5 & 0.1141 & No \\
\hline
\textbf{\#object} & \textbf{\#where} & \textbf{84.5} & \textbf{0.0001} & \textbf{Yes} \\
\hline
\#object & \#other & 21.5 & 0.0752 & No \\
\hline
\#where & \#other & 48.0 & 0.6815 & No \\
\hline
\end{tabular}
\caption{Results of the Mann-Whitney U Test for comparing \texttt{main-captions} dimensions.}
\label{tab:u_test_caption}
\end{table*}

\begin{table*}[h]
\centering
\begin{tabular}{|c|c|c|c|c|}
\hline
\textbf{Dimension1} & \textbf{Dimension2} & \textbf{U-statistic} & \textbf{P-value} & \textbf{Statistical Significance (0.05)} \\
\hline
\#action & \#subject & 23.5 & 0.5114 & No \\ \hline
\#action & \#object & 20.5 & 0.1570 & No \\ \hline
\#action & \#where & 14.0 & 0.1368 & No \\ \hline
\#action & \#other & 19.0 & 0.2526 & No \\ \hline
\#subject & \#object & 43.0 & 0.2747 & No \\ \hline
\#subject & \#where & 31.0 & 0.2682 & No \\ \hline
\#subject & \#other & 36.0 & 0.3049 & No \\ \hline
\#object & \#where & 49.5 & 0.7755 & No \\ \hline
\#object & \#other & 58.5 & 0.9472 & No \\ \hline
\#where & \#other & 47.5 & 0.8698 & No \\ \hline
\end{tabular}
\caption{Results of the Mann-Whitney U Test for comparing \texttt{main-news} dimensions.}
\label{tab:u_test_news}
\end{table*}

\begin{table*}[h]
\centering
\begin{tabular}{|c|c|c|c|c|}
\hline
\textbf{Dimension1} & \textbf{Dimension2} & \textbf{Statistic} & \textbf{P-value} & \textbf{Statistical Significance (0.05)} \\
\hline

\#action & \#subject & 0.0 & 0.9566 & No \\ \hline
\#action & \#object & 1.5 & 0.2269 & No \\ \hline
\#action & \#where & 0.0 & 0.9599 & No \\ \hline
\#action & \#other & 0.0 & 0.9543 & No \\ \hline
\#subject & \#object & 1.5 & 0.2234 & No \\ \hline
\#subject & \#where & 0.0 & 0.9935 & No \\ \hline
\#subject & \#other & 0.0 & 0.9291 & No \\ \hline
\#object & \#where & 1.4 & 0.2443 & No \\ \hline
\#object & \#other & 0.6 & 0.4614 & No \\ \hline
\#where & \#other & 0.0 & 0.9378 & No \\ \hline
\end{tabular}
\caption{Results of the Levene's Test for comparing \texttt{main-captions} dimensions.}
\label{tab:levene_test_caption}
\end{table*}

\begin{table*}[h]
\centering
\begin{tabular}{|c|c|c|c|c|}
\hline
\textbf{Dimension1} & \textbf{Dimension2} & \textbf{Statistic} & \textbf{P-value} & \textbf{Statistical Significance (0.05)} \\
\hline

\#action & \#subject & 2.0 & 0.1793 & No \\ \hline
\#action & \#object & 3.1 & 0.0986 & No \\ \hline
\#action & \#where & 3.4 & 0.0868 & No \\ \hline
\#action & \#other & 3.2 & 0.0967 & No \\ \hline
\#subject & \#object & 0.1 & 0.7885 & No \\ \hline
\#subject & \#where & 0.7 & 0.4097 & No \\ \hline
\#subject & \#other & 1.6 & 0.2186 & No \\ \hline
\#object & \#where & 0.5 & 0.5006 & No \\ \hline
\#object & \#other & 1.5 & 0.2343 & No \\ \hline
\#where & \#other & 0.4 & 0.5562 & No \\ \hline

\end{tabular}
\caption{Results of the Levene's Test for comparing \texttt{main-news} dimensions.}
\label{tab:levene_test_news}
\end{table*}

\subsubsection{Ground Score Rules}
\label{app:experiment_cog_diff_rules}

The ground score rules for STS-B and STS 2012-2016 are provided below, following the original papers \citep{Cer2017SemEval2017T1, Agirre2013SEM2S}.

\begin{itemize}
    \item \texttt{Score 5}: The two sentences are completely equivalent, as they mean the same thing.
    
    \item \texttt{Score 4}: The two sentences are mostly equivalent, but some unimportant details differ.
    
    \item \texttt{Score 3}: The two sentences are roughly equivalent, but some important information differs/missing.
    
    \item \texttt{Score 2}: The two sentences are not equivalent, but share some details.

    \item \texttt{Score 1}: The two sentences are not equivalent, but are on the same topic.
    
    \item \texttt{Score 0}: The two sentences are completely dissimilar.
    
\end{itemize}

\onecolumn

\onecolumn
\begin{center}
\begin{longtable}{|c|p{4cm}|p{4cm}|p{1.3cm}|p{1cm}|}
\caption{Sentence pair examples in \texttt{main-captions} with only one different dimension.}\label{tab:exp3_caption_examples} \\
\hline

\multicolumn{1}{|c|}{\textbf{Dimension}} & \multicolumn{1}{c|}{\textbf{Sentence1}} & \multicolumn{1}{c|}{\textbf{Sentence2}} & \multicolumn{1}{c|}{\textbf{Ground Score}} & \multicolumn{1}{c|}{\textbf{Data Source}} \\
\hline
\endfirsthead

\multicolumn{5}{c}%
{\tablename\ \thetable\ -- \textit{Continued from previous page}} \\
\hline
\multicolumn{1}{|c|}{\textbf{Dimension}} & \multicolumn{1}{c|}{\textbf{Sentence1}} & \multicolumn{1}{c|}{\textbf{Sentence2}} & \multicolumn{1}{c|}{\textbf{Ground Score}} & \multicolumn{1}{c|}{\textbf{Data Source}} \\
\hline
\endhead

\hline \multicolumn{5}{|r|}{\textit{Continued on next page}} \\ \hline
\endfoot

\hline
\endlastfoot

\texttt{\#action}  & A man is cycling & A man is talking & 0.6 & STS-B \\ \hline
\texttt{\#action}  & A man is speaking & A man is cooking & 0.8 & STS-B\\ \hline
\texttt{\#action}  & A man is praying & A man is dancing & 0.75 & STS-B\\ \hline
\texttt{\#action}  & A man is dancing & A man is thinking & 1.2 & STS-B\\ \hline
\texttt{\#action}  & people walk home & People waiting & 1.6 & STS-B\\ \hline
\texttt{\#action}  & A man is speaking & A man is spitting & 0.636 & STS-B\\ \hline
\texttt{\#action}  & A man is dancing & A man is singing & 1.25 & STS-B\\ \hline
\texttt{\#action}  & A man is running & A man is singing & 1.25 & STS-B\\ \hline
\texttt{\#action}  & A man is speaking & A man is dancing & 1.2 & STS-B\\ \hline
\texttt{\#action}  & A man is dancing & A man is speaking & 1.2 & STS-B\\ \hline
\texttt{\#action}  & Two dogs fighting in the snow & Two dogs standing in the snow & 2.4 & STS-B\\ \hline
\texttt{\#action}  & two dogs running in the snow & A dog laying in the snow & 2.2 & STS-B\\ \hline
\texttt{\#action}  & The people are leaving the airplane & The people are entering the plane & 2.2 & STS-B\\ \hline
\texttt{\#action}  & Two men are sitting in the room & Two men are standing in a room & 3.0 & STS-B\\ \hline
\texttt{\#action}  & A woman is writing & A woman is swimming & 0.5 & STS-B\\ \hline
\texttt{\#action}  & A man is levitating & A man is talking & 0.8 & STS-B\\ \hline
\texttt{\#action}  & A woman is slicing up some meat & A woman is breading some meat & 2.25 & STS-B\\ \hline
\texttt{\#action}  & A man is praying & A man is running & 0.727 & STS-B\\ \hline
\texttt{\#action}  & A man is singing & A man is dancing & 0.5 & STS-B\\ \hline
\texttt{\#action}  & A man is running & A man is mooing & 1.0 & STS-B\\ \hline
\texttt{\#action}  & A man is dangling a mouse near a snake & A man is feeding a mice to a snake & 3.2 & STS-B\\ \hline
\texttt{\#action}  & A animal is eating & The animal is hopping & 0.4 & STS-B\\ \hline
\texttt{\#action}  & A group of people eat at a table outside & Group of elderly people sitting around a table & 2.8 & STS-B\\ \hline
\texttt{\#action}  & A man and a woman laughing & A man and a woman kiss & 1.2 & STS-B\\ \hline

\texttt{\#subject}  & The ballerina is dancing & A man is dancing & 1.75 & STS-B \\ \hline
\texttt{\#subject} & Men are playing soccer & Two teams play soccer & 3.0 & STS-B \\ \hline
\texttt{\#subject} & A slow lori walks around & A animal is walking around & 2.8 & STS-B \\ \hline
\texttt{\#subject} & The lamb is looking at the camera & A cat looking at the camera & 0.8 & STS-B \\ \hline
\texttt{\#subject} & A skateboarder jumps off the stairs & A dog jumps off the stairs & 0.8 & STS-B \\ \hline
\texttt{\#subject} & Raccoons are eating & A man is eating & 1.5 & STS-B \\ \hline
\texttt{\#subject} & A Woman is eating & A animal is eating & 1.6 & STS-B \\ \hline
\texttt{\#subject} & A boy and a girl is dancing in the rain & A man and woman is dancing in the rain & 2.6 & STS-B \\ \hline
\texttt{\#subject} & A black and white cow standing in a grassy field & A blue jay standing in a grassy field & 0.6 & STS-B \\ \hline
\texttt{\#subject} & Old green bottle sitting on a table & Three men in suits sitting at a table & 0.4 & STS-B \\ \hline
\texttt{\#subject} & Long - haired black dog standing in grassy field & A blue jay standing in a grassy field & 1.4 & STS-B \\ \hline
\texttt{\#subject} & Three dogs racing on a dirt track & Cars racing on a dirt track & 1.4 & STS-B \\ \hline

\texttt{\#object} & A woman is cutting onions & A woman is cutting tofu & 1.8 & STS-B\\ \hline
\texttt{\#object} & A man is slicing a tomato & A man is slicing a bun & 2.0 & STS-B\\ \hline
\texttt{\#object} & A woman is cutting tofu & A woman is cutting an onion & 2.4 & STS-B\\ \hline
\texttt{\#object} & A woman is cutting some fish & A woman is cutting tofu & 2.2 & STS-B\\ \hline
\texttt{\#object} & A woman is slicing ginger & A woman is cutting potatoes & 1.6 & STS-B\\ \hline
\texttt{\#object} & A group of kids are having a jumping contest & A group of kids are having a sleepover & 1.2 & STS-B\\ \hline
\texttt{\#object} & The dog is chasing the geese & One dog is chasing the other & 1.6 & STS-B\\ \hline
\texttt{\#object} & A man is playing soccer & A man is playing flute & 1.0 & STS-B\\ \hline
\texttt{\#object} & A girl is riding a horse & A girl is riding a bicycle & 1.917 & STS-B\\ \hline
\texttt{\#object} & A man is playing a basketball & A man is playing a piano & 1.4 & STS-B\\ \hline
\texttt{\#object} & A man is playing an electronic keyboard & A man is playing a flute & 1.2 & STS-B\\ \hline
\texttt{\#object} & A guy is playing hackysack & A man is playing a key - board & 1.0 & STS-B\\ \hline
\texttt{\#object} & A woman is slicing a potato & A woman is slicing carrot & 2.5 & STS-B\\ \hline
\texttt{\#object} & A man is cutting up a potato & A man is cutting up carrots & 2.375 & STS-B\\ \hline
\texttt{\#object} & The gate is blue & The gate is yellow & 1.6 \\ \hline
\texttt{\#object} & A man is slicing a bun & A man is slicing an onion & 2.4 & STS-B\\ \hline
\texttt{\#object} & A man is cutting up a potato & A man is cutting up carrots & 2.375 & STS-B\\ \hline
\texttt{\#object} & A woman is peeling a potato & A woman is peeling an apple & 2.0 & STS-B\\ \hline
\texttt{\#object} & The men are playing cricket & The men are playing basketball & 2.2 & STS-B\\ \hline
\texttt{\#object} & A woman is doing weight exercises & A woman is doing her hair & 0.5 & STS-B\\ \hline

\texttt{\#object} & The man is slicing a fish open & A man is slicing a potato & 2.2 & STS-B\\ \hline
\texttt{\#object} & A woman is cutting an apple & A woman is cutting potato & 1.8 & STS-B\\ \hline
\texttt{\#object} & A woman is cutting some flowers & A woman is cutting broccoli & 1.0 & STS-B\\ \hline
\texttt{\#object} & A woman is cutting potatoes & A woman is slicing carrots & 1.2 & STS-B\\ \hline
\texttt{\#object} & A woman is chopping garlic & A woman slices a fish & 1.0 & STS-B\\ \hline

\texttt{\#where} & A yellow bird is eating fruit on a wire grate & A yellow bird eating fruit on a bird feeder & 2.8 & STS-B \\
\hline
\texttt{\#where} & Two dogs play in the grass & Two dogs playing in the snow & 2.8 & STS-B \\
\hline
\texttt{\#where} & A brown and white dog is running through the snow & a brown and white dog is running on the grass & 2.4 & STS-B \\
\hline
\texttt{\#where} & Three children playing in snow & Three children playing in hay & 2.4 & STS-B \\
\hline
\texttt{\#where} & Two dogs playing in snow & Two dogs playing in grass & 2.4 & STS-B \\
\hline
\texttt{\#where} & A jockey riding a horse in a pen & A jockey rides a horse at a gallop & 3.6 & STS-B \\
\hline
\texttt{\#where} & A red and white plane flying on a sunny day & Red and white plane flying through the air & 4.2 & STS-B \\
\hline
\texttt{\#where} & Two women sitting in lawn chairs & Two women are sitting in a cafe & 2.6 & STS-B \\
\hline
\texttt{\#where} & A white cat laying on an office chair & A white cat laying on a sheet & 2.4 & STS-B \\
\hline
\texttt{\#where} & Two girls walking in the ocean & two girls walking in the street & 1.8 & STS-B \\
\hline
\texttt{\#where} & Three dogs are playing in the white snow & Two dogs are playing in the grass & 1.6 & STS-B \\
\hline
\texttt{\#where} & A man and woman are driving down the street in a jeep & A man and woman are driving down the road in an open air vehicle & 4.0 & STS-B \\
\hline
\texttt{\#where} & Two men are fighting in a cow pasture & Two men are fighting in a cattle pen & 4.0 & STS-B \\
\hline
\texttt{\#where} & People sitting on the porch & People sitting on a couch & 1.4 & STS-B \\
\hline
\texttt{\#where} & A white jeep parked in front of a store & A white Jeep parked on a street & 3.5 & STS-B \\
\hline
\texttt{\#where} & The two dogs are running through the grass & Three dogs run in the snow & 2.2 & STS-B \\
\hline
\texttt{\#where} & A dog runs through the grass & A dog runs through the snow & 2.2 & STS-B \\
\hline
\texttt{\#where} & A black dog running in the snow & A black dog running on a beach & 1.8 & STS-B \\
\hline
\texttt{\#where} & a man walks two dogs on leashes down the street & A man walks two dogs in the city & 4.2 & STS-B \\
\hline
\texttt{\#where} & A dog is running in the snow & Two dogs running in the dirt & 2.2 & STS-B \\
\hline
\texttt{\#where} & A brown dog is running through green grass & A brown dog is running though a river & 2.4 & STS-B \\
\hline

\texttt{\#other} & The dog is running with food in his mouth & The dog is running with a yellow ball in his mouth & 2.2 & STS-B\\ \hline
\texttt{\#other}  & A baby tiger is playing with a ball & A baby is playing with a doll & 1.6 & STS-16\\ \hline
\texttt{\#other}  & A man is running with a bus &  A man runs with a truck & 2.5  & STS-16\\ \hline
\texttt{\#other}  & A cat is rubbing against baby's face & a cat is rubbing against a baby & 3.8 & STS-12\\ \hline

\end{longtable}
\end{center}


\begin{center}
\begin{longtable}{|c|p{4cm}|p{4cm}|p{1.3cm}|p{1cm}|}
\caption{Sentence pair examples in \texttt{main-news} with only one different dimension.}\label{tab:exp3_news_examples} \\
\hline

\multicolumn{1}{|c|}{\textbf{Dimension}} & \multicolumn{1}{c|}{\textbf{Sentence1}} & \multicolumn{1}{c|}{\textbf{Sentence2}} & \multicolumn{1}{c|}{\textbf{Ground Score}} & \multicolumn{1}{c|}{\textbf{Data Source}} \\
\hline
\endfirsthead

\multicolumn{5}{c}%
{\tablename\ \thetable\ -- \textit{Continued from previous page}} \\
\hline
\multicolumn{1}{|c|}{\textbf{Dimension}} & \multicolumn{1}{c|}{\textbf{Sentence1}} & \multicolumn{1}{c|}{\textbf{Sentence2}} & \multicolumn{1}{c|}{\textbf{Ground Score}} & \multicolumn{1}{c|}{\textbf{Data Source}} \\
\hline
\endhead

\hline \multicolumn{5}{|r|}{\textit{Continued on next page}} \\ \hline
\endfoot

\hline
\endlastfoot

\texttt{\#action} & Saudi Arabia gets a seat at the UN Security Council & Saudi Arabia rejects seat on UN Security Council & 2.8 & STS-B\\ \hline
\texttt{\#action} & Romney wins Florida Republican primary & Romney eyes US Republican primary endgame & 1.8 & STS-B\\ \hline
\texttt{\#action} & Hong Kong stocks close down 0.28 \% & Hong Kong stocks open 0.62 pct higher & 1.6 & STS-B\\ \hline
\texttt{\#action} & French train derails south of Paris & French train passengers tell of crash ordeal & 1.2 & STS-B\\ \hline
\texttt{\#action} & Indian stocks open lower & Indian stocks close lower & 1.8 & STS-B\\ \hline

\texttt{\#action} & FAA continues ban on US flights to Tel Aviv & FAA lifts ban on U.S. flights to Tel Aviv & 1.8 & STS-B\\ \hline

\texttt{\#subject} & It has a margin of error of plus or minus 4 percentage points & That poll had 712 likely voters and sampling error of plus or minus 3.7 percentage points & 2.5  & STS-B\\ \hline
\texttt{\#subject} & 20 killed in bomb attack at Pakistani base & 8 soldiers killed in bomb attack in NW Pakistan & 2.0  & STS-B\\ \hline
\texttt{\#subject} & Pak religious body endorses underage marriage & CII endorses underage marriage & 3.0  & STS-B\\ \hline
\texttt{\#subject} & At least 15 killed in Nigeria church attack & At least 12 killed in Nigeria church bombing & 2.6  & STS-B\\ \hline
\texttt{\#subject} & Brooks pleads not guilty to hacking charges & Dave Lee Travis pleads not guilty to all charges & 1.4  & STS-B\\ \hline
\texttt{\#subject} & tirana is the capital of & abuja is the capital of & 1.0  & STS-B\\ \hline
\texttt{\#subject} & Santorum's 3 - year - old daughter hospitalized & 'Kony 2012' director hospitalized & 0.6  & STS-B\\ \hline
\texttt{\#subject} & 7 killed in attacks in Iraq & 27 killed in attacks across Iraq & 2.0  & STS-B\\ \hline
\texttt{\#subject} & Six Australians killed in Laos plane crash & Dozens killed in Laos plane crash & 3.0  & STS-B\\ \hline
\texttt{\#subject} & Philip leaves hospital after 11 days & Mandela leaves hospital after 10 days & 1.8  & STS-B\\ \hline

\texttt{\#object} & US drone strike kills 11 in Pakistan & US drone kills 16 in Pakistan & 3.2 & STS-B\\\hline
\texttt{\#object} & Suicide attack kills eight in Baghdad & Suicide attacks kill 24 people in Baghdad & 2.4 & STS-B\\\hline
\texttt{\#object} & US drone strike 'kills 4 militants in Pakistan' & US drone strike kills 10 in Pakistan & 3.8 & STS-B\\\hline
\texttt{\#object} & US drone strike kills four in North Waziristan & U.S. drone strike kills 10 in northwest Pakistan: officials & 1.8 & STS-B\\\hline
\texttt{\#object} & Indonesian president to visit UK & Indonesian president to visit Australia & 1.4 & STS-B\\\hline
\texttt{\#object} & 6.0-magnitude quake hits northern Italy: USGS & 5.9-magnitude quake hits Sunda Strait, Indonesia: USGS & 1.6 & STS-B\\\hline
\texttt{\#object} & 6.3-magnitude earthquake hits Taiwan & 7.7-magnitude earthquake hits SW Pakistan & 1.2 & STS-B\\\hline
\texttt{\#object} & Roadside bombs kill 5 in Afghanistan & Roadside bomb kills 3 policemen in Afghanistan & 3.4 & STS-B\\\hline
\texttt{\#object} & US drone strike kills three in northwest Pakistan & US drone strike kills seven in North Waziristan & 2.4 & STS-B\\\hline
\texttt{\#object} & Iraq violence kills 11 & Iraq violence kills seven & 2.2 \\ \hline
\texttt{\#object} & Strong earthquake in western China kills 47 people & Strong earthquake in western China kills at least 75 & 3.0 & STS-B\\\hline
\texttt{\#object} & Bomb attacks kill 20 in Baghdad's Christian areas & Bombs Kill 35 in Baghdad Christian Area & 3.0 & STS-B\\\hline

\texttt{\#where} & Pakistan blocks Twitter over anti - Islamic material & Pakistan blocks Twitter over ' blasphemy ' & 3.8 & STS-B\\ \hline
\texttt{\#where} & Shenzhen stock indices close higher Monday & Shenzhen stock indices close lower -- Oct. 31 & 1.8 & STS-B\\ \hline
\texttt{\#where} & Two NATO soldiers killed in Afghanistan & NATO soldier killed in Afghan attack & 4.0 & STS-B\\ \hline
\texttt{\#where} & NATO Soldier Killed In Afghan Attack & NATO soldier killed in S. Afghanistan & 3.8 & STS-B\\ \hline
\texttt{\#where} & Two NATO soldiers killed in Afghanistan & NATO Soldier Killed in Afghan Blast & 3.6 & STS-B\\ \hline
\texttt{\#where} & ' Scores of bodies ' found in Syria & Eight more bodies found on ship & 1.4 & STS-B\\ \hline
\texttt{\#where} & Musharraf arrested in Lal Masjid case & Musharraf arrested in Pakistan & 3.0 & STS-B\\ \hline
\texttt{\#where} & Hushen 300 Index closes higher -- Oct. 14 & Hushen 300 Index closes lower -- March 12 & 2.0 & STS-B\\ \hline
\texttt{\#where} & Seven peacekeepers killed in Sudan's Darfur & Peacekeeper killed in Abyei clash & 1.6 & STS-B\\ \hline
\texttt{\#where} & Powerful 7.6 quake strikes off Solomons & Powerful 6.9 quake strikes off California coast & 2.2 & STS-B\\ \hline

\texttt{\#other} & 5 Things to Know About the Sochi Olympics & 7 Things to Know About Ethanol & 0.0 & STS-B\\ \hline
\texttt{\#other}  & van der merwe sentenced geiges to a total of 13 years imprisonment & van der merwe suspended geiges' sentence to 5 years imprisonment & 2.6 & STS-B\\ \hline
\texttt{\#other}  & Death toll in Nigeria police attack rises to 30 & Death toll in Kenya bus attack rises to six & 1.4 \\ \hline
\texttt{\#other}  & Residents return to Texas blast site & Residents return to Fallujah & 0.0 & STS-B\\ \hline
\texttt{\#other}  & China yuan strengthens to 6.2689 against USD & China yuan strengthens to new high against USD & 3.0 & STS-B\\ \hline
\texttt{\#other}  & Death toll from Philippine earthquake rises to 185 & Death toll from Philippines quake rises to 144 & 3.6 & STS-B\\ \hline
\texttt{\#other}  & Stars pay tribute to Cory Monteith & Stars pay tribute to James Garner & 2.2 & STS-B\\ \hline
\texttt{\#other}  & Death toll from Egypt protests rises to 49 & Death toll from Egypt violence rises to 638 & 3.2 & STS-B\\ \hline
\texttt{\#other}  & Newark mayor saves neighbor from fire & Newark mayor rescues neighbor from burning house & 4.2 & STS-B\\ \hline
\texttt{\#other}  & 10 Things to Know for Wednesday & 10 Things to Know for Today & 3.0 & STS-B\\ \hline

\end{longtable}
\end{center}


\twocolumn

\end{document}